\documentclass[acmtog, screen]{acmart}

\makeatletter
\renewcommand\@formatdoi[1]{\ignorespaces}
\makeatother

\renewcommand\footnotetextcopyrightpermission[1]{} % removes footnote with conference information in first column
\usepackage{booktabs} %
\usepackage{array,longtable}
\usepackage{multirow}
\usepackage{adjustbox}
\usepackage[ruled]{algorithm2e} %

\acmJournal{JOCCH}
\acmVolume{X}
\acmNumber{X}
\acmArticle{XX}
\acmYear{2019}
\acmMonth{0}
\acmArticleSeq{00}

\setcopyright{rightsretained}

\acmDOI{}

\begin{document}

\title{Egocentric Visitors Localization in Cultural Sites}

\author{Francesco Ragusa}

\affiliation{%
	\institution{DMI - IPLab, Universit\`a degli Studi di Catania}
	\streetaddress{}
	\city{Catania}
	\state{CT}
	\country{Italia}}
\email{francesco.ragusa@unict.it}

\author{Antonino Furnari}

\affiliation{%
	\institution{DMI - IPLab, Universit\`a degli Studi di Catania}
	\streetaddress{}
	\city{Catania}
	\state{CT}
	\country{Italia}}
\email{furnari@dmi.unict.it}

\author{Sebastiano Battiato}

\affiliation{%
	\institution{DMI - IPLab, Universit\`a degli Studi di Catania}
	\streetaddress{}
	\city{Catania}
	\state{CT}
	\country{Italia}}
\email{battiato@dmi.unict.it}

\author{Giovanni Signorello}

\affiliation{%
	\institution{CUTGANA, Universit\`a degli Studi di Catania}
	\streetaddress{}
	\city{Catania}
	\state{CT}
	\country{Italia}}
\email{g.signorello@unict.it}

\author{Giovanni Maria Farinella}
\authornote{This is the corresponding author}

\affiliation{%
	\institution{DMI - IPLab \& CUTGANA, Universit\`a degli Studi di Catania}
	\streetaddress{}
	\city{Catania}
	\state{CT}
	\country{Italia}}
\email{gfarinella@dmi.unict.it}

\renewcommand\shortauthors{F. Ragusa, A. Furnari, S. Battiato, G. Signorello, G. M. Farinella}

\begin{abstract}
We consider the problem of localizing visitors in a cultural site from egocentric (first person) images. Localization information can be useful both to assist the user during his visit (e.g., by suggesting where to go and what to see next) and to provide behavioral information to the manager of the cultural site (e.g., how much time has been spent by visitors at a given location? What has been liked most?). To tackle the problem, we collected a large dataset of egocentric videos using two cameras: a head-mounted HoloLens device and a chest-mounted GoPro. 
Each frame has been labeled according to the location of the visitor and to what he was looking at. 
The dataset is freely available in order to encourage research in this domain. The dataset is complemented with baseline experiments performed considering a state-of-the-art method for location-based temporal segmentation of egocentric videos. 
Experiments show that compelling results can be achieved to extract useful information for both the visitor and the site-manager.
\end{abstract}




\maketitle

\section{Introduction}
\label{sec:introduction}
Cultural sites receive lots of visitors every day. To improve the
fruition of cultural objects, a site manager should be able to assist the users
during their visit by providing additional information and suggesting what to
see next, as well as to gather information to understand the behavior of the
visitors (e.g., what has been liked most) in order to improve the suggested
visit paths or the placement of artworks. Traditional ways to achieve such goals
include the employment of professional guides, the installation of informative
panels, the distribution of printed material to the users (e.g., maps and
descriptions) and the collection of visitors' opinions through surveys. When the
number of visitors grows large, the aforementioned traditional tools tend to
become less effective, which motivates the employment of automated technologies.
In order to assist visitors in a scalable and interactive way, site managers
have employed technologies aimed at providing complementary information on the
cultural objects on demand. An example of such technologies are audio guides,
which allow to obtain spoken information about a point of interest by dialling
the appropriate number on the device. Similarly, the use of tablets or
smartphones allows to obtain audio-visual complementary information of an
observed object of the cultural site by interacting with a touch interface
(e.g., inserting the number of the cultural object of interest) or by taking a
picture of a QR Code. Although effective in some cases, the
aforementioned technologies are very limited by the following factors: 
\begin{itemize}
	\item they require the active intervention of the visitor, who needs to specify
	the correct number or to take a picture of the right QR Code; 
	\item they require the site manager to install informative panels reporting the
	number or QR Code corresponding to a given cultural object (which is sometimes not
	possible due to the nature of the site).

\end{itemize}
Moreover, traditional systems are unable to acquire any information useful to
understand the visitor's habits or interests. To gather information about the visitors (i.e., what they see and where they are) in
an automated way, past works have employed fixed cameras and classic ``third
person vision'' algorithms to detect, track, count people and estimate their
gaze~\cite{bartoli2015museumvisitors}. However, systems based on third person
vision are capped by several limitations: 1) fixed cameras need to be installed
in the cultural site and this is not always possible, 2) the fixed viewpoint of third person cameras makes it
difficult to estimate what the visitors are looking at (e.g., ambiguity on estimation of what people see), 3) fixed cameras are
easily affected by occlusion and people re-identification problems (e.g., difficulties to follow a person from a room to another), 3) the
system has to work for several visitors at a time, making it difficult to
profile them and to adapt its functioning to their specific needs (e.g., personal recommendation). Moreover,
systems based on third person vision cannot easily communicate to the visitor in
order to ``augment his visit'' providing information on the observed
cultural object or by recommending what to see next.

Ideally, we would like to provide the user with an unobtrusive wearable device
capable of addressing both tasks: augmenting the visit and inferring behavioral
information about the visitors. We would like to note that wearable devices are particularly suited to solve this kind of tasks as they are naturally worn and carried by the visitor. Moreover, wearable systems do not require the explicit intervention of the visitor to deliver services such as localization, augmented reality and recommendation. The device should be aware of the current
location of the visitor and capable to infer what he is looking at, and,
ultimately, his behavior (e.g., what has already been seen? for how long?).
Such a system would allow to provide automatic assistance to the visitor by
showing him the current location, guiding him to a given area of the
site, giving information about the currently observed cultural object, keeping
track of what has been already seen and for how long, and suggesting what is yet
to be seen by the visitor. Equipping multiple visitors with an egocentric vision device,
it would be possible to track a profile of the different visitors in order to
provide: 1) recommendations on what to see in the cultural site based on what
has already been seen/liked (e.g, considering how much time has been spent at a
given location or for how long the user has observed a cultural object), 2)
statistics on the behaviors of the visitors within the site. Such statistics
could be of great use by the site manager to improve the services offered by the
cultural site and to facilitate the fruition of the cultural site.

\begin{figure*}
	\centering
	\includegraphics[width=1.0\linewidth]{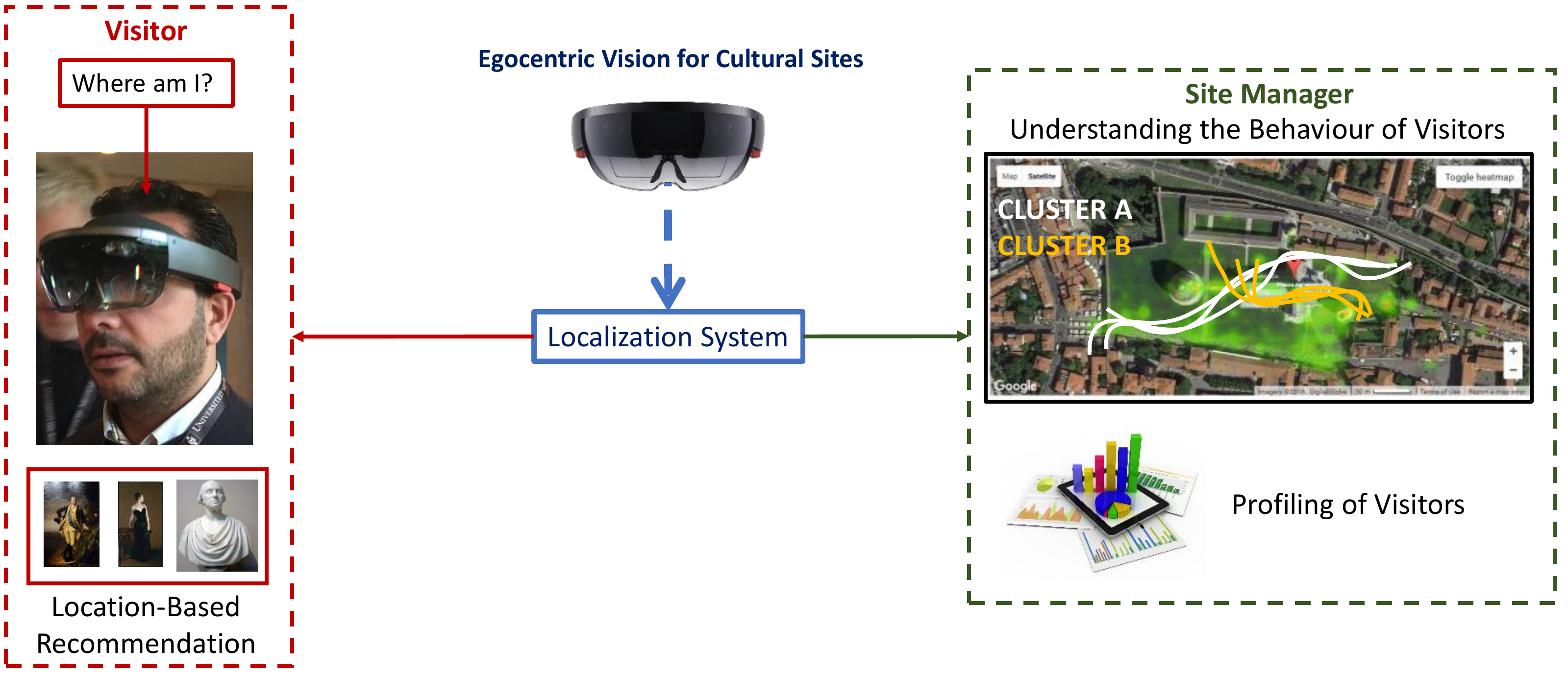}	
	\qquad%
	\caption{A diagram of a system which uses egocentric visitor localization to	provide assistance to the user and augment his visit (left) and to provide	useful information to the site manager (right).}
	\label{fig:concept}
\end{figure*}

As investigated by other
authors~\cite{cucchiara2014visions,colace2014context,taverriti2016real,seidenari2017deep},
wearable devices equipped with a camera such as smart glasses (e.g., Google
Glass, Microsoft HoloLens and Magic Leap) offer interesting opportunities to
develop the aforementioned technologies and services for visitors and site
managers. Wearable glasses equipped with mixed reality visualization systems
(i.e., capable of displaying virtual elements on images coming from the real
world) such as Microsoft HoloLens and Magic Leap allow to provide information to
the visitor in a natural way, for example by showing a 3D reconstruction of a
cultural object or by showing virtual textual information next to a work of art.
In particular, a wearable system should be able to carry out at least the
following tasks: 1) localizing the visitor at any moment of the visit, 2)
recognizing the cultural objects observed by the visitor, 3) estimating the
visitor's attention, 4) profiling the user, 5) recommending what to see next.

In this work, we address egocentric visitor localization, the first step towards the construction of a wearable
system capable of assisting the visitor of a cultural site and collect
information useful for the site management. In particular, we concentrate on the
problem of room-based localization of visitors in cultural sites from egocentric
visual data. As it is depicted in~\figurename~\ref{fig:concept}, egocentric localization of visitors already enables different applications providing services to both the visitor and the site manager. In particular, localization allows to implement a ``where am I'' service, to provide the user which his own location in the cultural site and a location-based recommendation system. By collecting and managing localization information over time, the site manager will be able to profile visitors and understand their behavior. 
To study the problem we collected and labeled a large dataset of
egocentric videos in the cultural site ``Monastero dei
Benedettini''\footnote{www.monasterodeibenedettini.it/en}, UNESCO World
Heritage Site, which is located in Catania, Italy. The dataset has been acquired
with two different devices and contains more than $4$ hours of video. Each frame
has been labeled according to the location in which the user was
located at the moment of data acquisition, as well as with the cultural object
observed by the subject (if any). The dataset is publicly available for research
purposes at \url{http://iplab.dmi.unict.it/VEDI/}.

We also report baseline
results for the the problem of room-based localization on the proposed dataset. Experimental results point out that useful information such as the time spent in each location
by visitors can be effectively obtained with egocentric systems.

The rest of the paper is organized as follows. In Section~\ref{sec:relatedwork} we discuss the related work. The dataset proposed in this study is presented in Section~\ref{sec:dataset}. Section~\ref{sec:method} revises the baseline approach for room-based localization of visitors in a cultural site. The experimental settings are reported in Section~\ref{sec:experiments}, while the results as well as a graphical interface to allow the analysis of the egocentric videos by the site manager are presented in Section~\ref{sec:results}. Section~\ref{sec:conclusion} concludes the paper.

\section{Related Work}
\label{sec:relatedwork}

The use of Computer Vision to improve the fruition of cultural objects has been
already investigated in past studies. Cucchiara and Del Bimbo discuss the use of
computer vision and wearable devices for augmented cultural experiences
in~\cite{cucchiara2014visions}. In~\cite{colace2014context} it is presented the
design for a system to provide context aware applications and assist tourists.
In~\cite{taverriti2016real} it is described a system to perform
real-time object classification and artwork recognition on a wearable device.
The system makes use of a Convolutional Neural Network (CNN) to perform object
classification and artwork classification. In~\cite{portaz2017fully} is discussed
an approach for egocentric image classification and object detection
based on Fully Convolutional Networks (FCN). The system is adapted to mobile
devices to implement an augmented audio-guide.
In~\cite{gallo2017exploiting}, it is proposed a method to exploit georeferenced
images publicly available on social media platforms to get insights on the
behavior of tourists. In~\cite{seidenari2017deep} is addressed the
problem of creating a smart audio guide that adapts to the actions and interests
of visitors. The system uses CNNs to perform object classification and
localization and is deployed on a NVIDIA Jetson TK1. In ~\cite{signorello2015exploring} is
investigated multimodal navigation of
multimedia contents for the fruition of protected natural areas.

Visitors' localization can be tackled outdoor using Global Positioning System
(GPS) devices. These systems, however, are not suitable to localize the user in
an indoor environment. Therefore, different Indoor Positioning
Systems (IPS) have been proposed through the years~\cite{curran2011evaluation}.
In order to retrieve accurate positions, these systems rely on devices such as
active badges~\cite{want1992active} and WiFi networks~\cite{gu2009survey}, which
need to be placed in the environments and hence become part of the
infrastructure. This operational way is not scalable since it requires the
installation of specific devices, which is expensive and not always feasible,
for instance, in the context of cultural heritage. Visual localization can be
used to overcome many of the considered challenges. For instance, previous works addressed visual landmark recognition with smartphones~\cite{amato2015fast,WEYAND20151,li2017visual}. In particular, the use of a
wearable cameras allows to localize the user without relying on specific
hardware installed in the cultural site. Visual localization can be performed at
different levels, according to the required localization precision and to the
amount of available training data. Three common levels of localization are scene
recognition~\cite{oliva2001modeling,zhou2014learning}, location
recognition~\cite{pentland1998visual,aoki1998recognizing,torralba2003context,furnari2018personal}
and 6-DOF camera pose
estimation~\cite{shotton2013scene,kendall2015posenet,sattler2017efficient}. Some
works also investigated the combination of classic localization based on
non-visual sensors (such as bluetooth) with computer
vision~\cite{ishihara2017inference,ishihara2017beacon}.

In this work, we concentrate on location recognition, since we want to be able
to recognize the environment (e.g., room) in which the visitor is located.
Location recognition is the ability to recognize when the user is moving in a
specific space at the instance level. In this case the egocentric (first person) vision system should
be able to understand if the user is in a given location. Such location can
either be a room (e.g., office 368 or exhibition room 3) or a personal space
(e.g., office desk). In order to setup a location recognition system, it is
usually necessary to acquire a moderate amount of visual data covering all the
locations visited by the user. Visual location awareness has been investigated
by different authors over the years. In~\cite{pentland1998visual} has been
addressed the recognition of basic tasks and locations related to the Patrol
game from egocentric videos in order to assist the user during the game. The
system was able to recognize the room in which the user was operating using
simple RGB color features. An Hidden Markov Model (HMM) was employed to enforce the temporal
smoothness of location predictions over time. In ~\cite{aoki1998recognizing}, it is proposed a system
to recognize personal locations from
egocentric video using the approaching trajectories observed by the wearable
camera. At training time the system built a dictionary of visual trajectories
(i.e., collections of images) captured when approaching each specific location.
At test time, the observed trajectory was matched to the dictionary in order to
detect the current location. In ~\cite{torralba2003context}, it has been designed
a context-based vision system for place and scene recognition. The system used
an holistic visual representation similar to GIST to detect the current location
at the instance level and recognize the scene category of previously unseen
environment. Other authors~\cite{xu2014wearable} proposed a way to provide
context-aware assistance for indoor navigation using a wearable system. When it
is not possible to acquire data for all the locations which might be visited by
the user, it is generally necessary to explicitly consider a rejection option,
as proposed in~\cite{furnari2018personal}.

Some authors proposed datasets to investigate different problems related to
cultural sites. For instance, In ~\cite{portaz2017fully,portaz2017construction}, it is proposed a dataset of images
acquired by using classic or head-mounted cameras. The dataset contains a small
number of images and is intended to address the problem of image search (e.g.,
recognizing a painting). In~\cite{bartoli2015museumvisitors}, is
presented a dataset acquired inside the National Museum of Bargello in Florence.
The dataset (acquired by 3 fixed IPcameras) is intended for pedestrian and group detection, gaze
estimation and behavior understanding.

To the best of our knowledge, this paper
introduces the first large scale public dataset acquired in a cultural site using
wearable cameras which is intended for egocentric visitor localization research purposes.
\begin{figure*}
	\centering
	\includegraphics[width=1.0\linewidth]{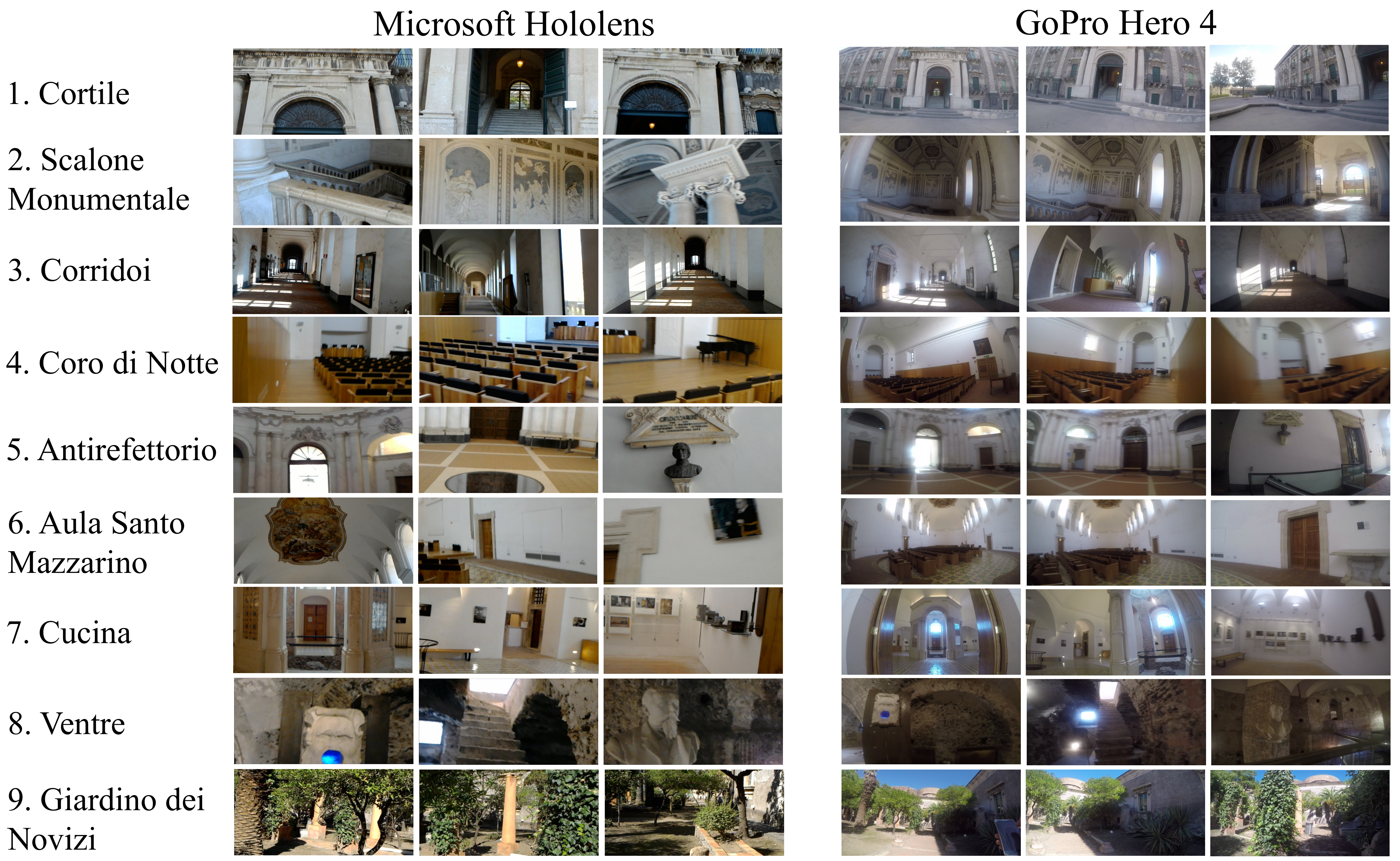}	
	\qquad%
	\caption{The figure shows some frames for each considered environment, acquired
		with Microsoft Hololens (left column) and GoPro Hero4 (right column) wearable
		devices.}
	\label{fig:environments}
\end{figure*}

\section{Dataset}
\label{sec:dataset}
We collected a large dataset of videos acquired in the \textit{Monastero dei
	Benedettini}, located in Catania, Italy.
The dataset has been acquired using two wearable devices: a
head-mounted Microsoft HoloLens and a chest-mounted GoPro Hero4. The considered
devices represent two popular choices for the placement of wearable cameras.
Indeed, chest-mounted devices generally allow to produce better quality images
due to the reduced egocentric motion with respect to head-mounted devices. On
the other side, head-mounted cameras allow to capture what the user is looking
at and hence they are better suited to model the attention of the user.
Moreover, head-mounted devices such as HoloLens allow for the use of augmented
reality, which can be useful to assist the visitor of a cultural site. The two
devices are also characterized by two different Field Of View (FOV). In
particular, the FOV of the HoloLens device is narrow-angle, while the FOV of the
GoPro device is wide-angle. This is shown in \figurename~\ref{fig:environments}, which reports some frames acquired with HoloLens along with the corresponding images acquired with GoPro.
Since we would like to assess which device is better suited to address the
localization problem, we use the two devices simultaneously during the data
acquisition procedure in order to collect two separate and compliant datasets: one
containing only data acquired with HoloLens device and the other one
containing only data acquired using the GoPro device. The videos captured by HoloLens device have a resolution equal to $1216 \times 684$ pixels and a
frame rate of $24$ fps, whereas the videos recorded with GoPro Hero~4 have a resolution
of $1280 \times 720$ pixels and a frame rate of $25$ fps.

\begin{figure*}
	\centering
	\includegraphics[width=0.85\linewidth]{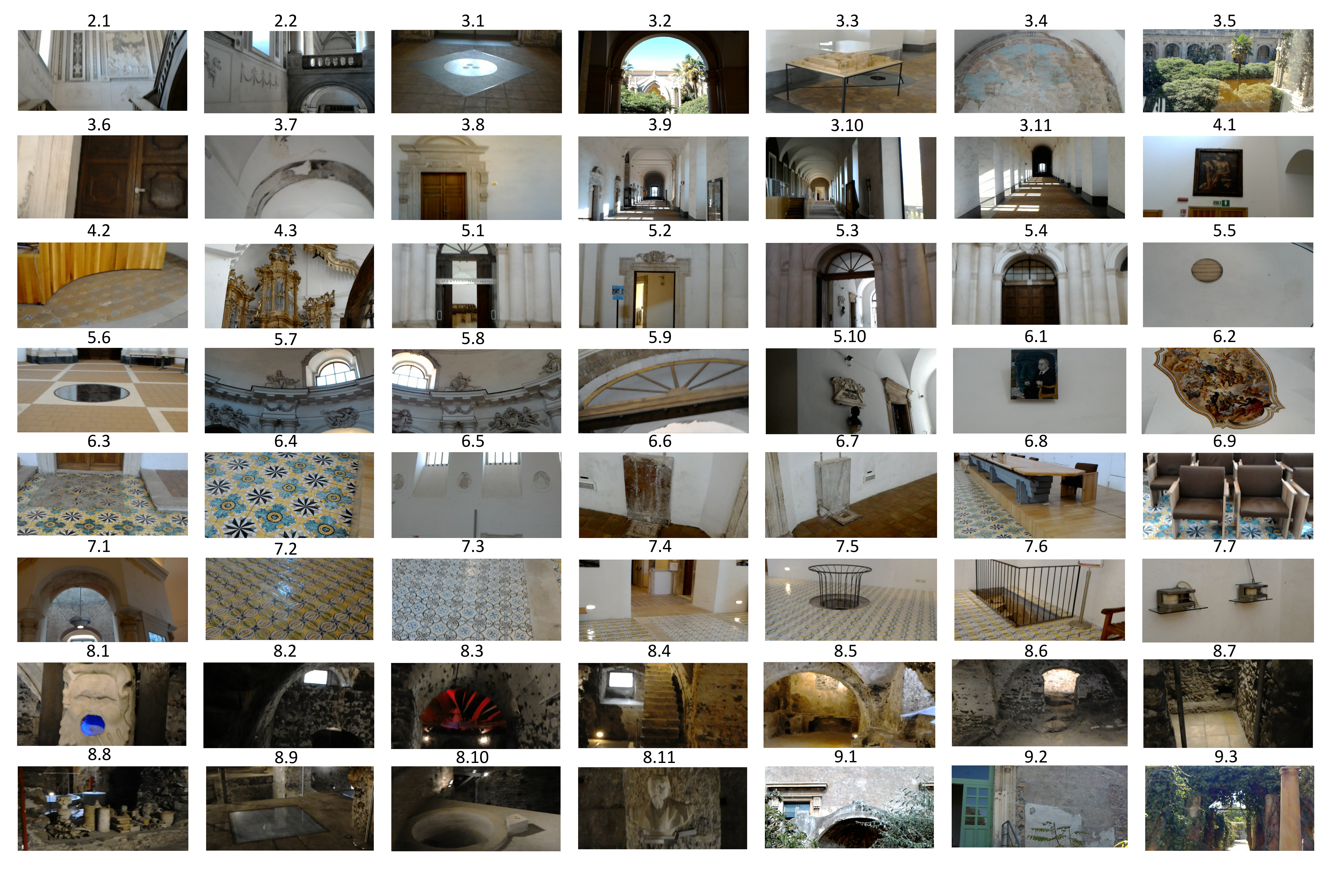}	
	\includegraphics[width=0.85\linewidth]{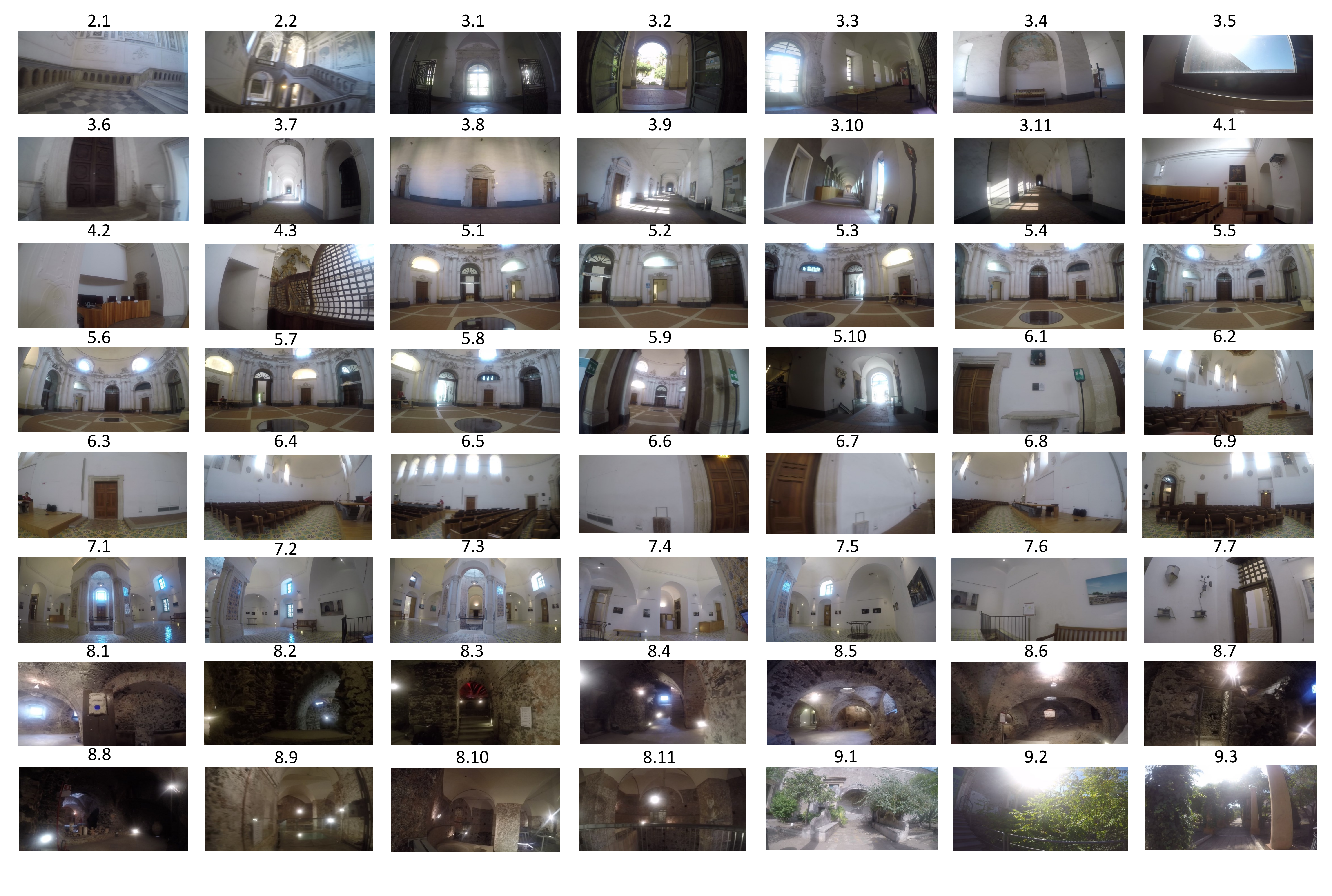}	
	\qquad%
	\caption{The figure shows a sample frame for each of the $57$ points of
		interest acquired with both Microsoft Hololens (top) and GoPro (bottom).}
	\label{fig:interest}
\end{figure*}

\begin{figure*}
	\centering
	
	{{\includegraphics[width=\linewidth]{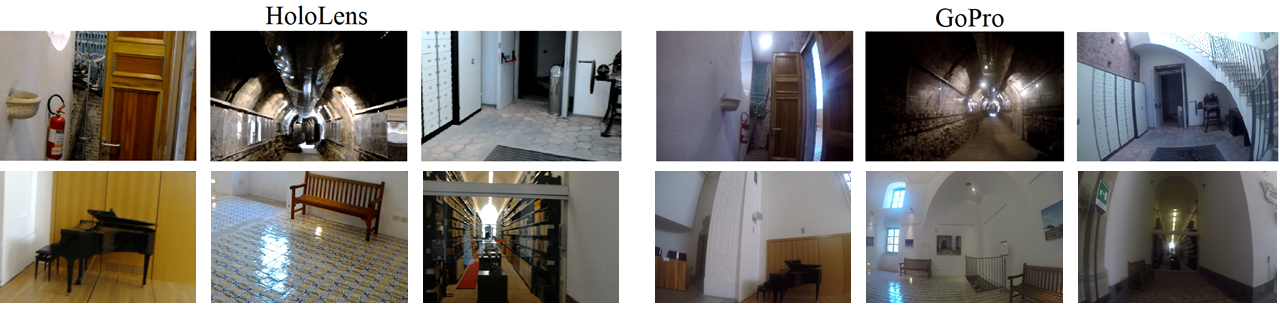}}}		
	\qquad%
	\caption{Example of frames belonging to the negative class, acquired with
		Microsoft Hololens (left) and GoPro (right).}
	\label{fig:negclass}
\end{figure*}

Each video frame has been labeled according to: 1) the location of the
visitor and 2) the ``point of interest'' (i.e. the cultural object)
observed by the visitor, if any. In both cases, a frame can be labeled as
belonging to a ``negative'' class, which denotes all visual information which is
not of interest. For example, a frame is labeled as negative when the visitor
is transiting through a corridor which has not been included in the training set
because is not a room (context) of interest for the visitors or when he is not looking at any
of the considered points of interest. We considered a total of $9$ environments
and $57$ points of interests. Each environment is identified by a number from
$1$ to $9$, while points of interest (i.e., cultural objects) are denoted by a code in the form $X.Y$
(e.g., $2.3$), where $X$ denotes the environment in which the points of interest
are located and $Y$ identify the point of interest.
\figurename~\ref{fig:environments} shows some representative samples for each of
the $9$ considered environments. \tablename~\ref{table:labels} shows the list of
the considered environments (left column) and the related points of interest
(right column). In the case of class \textit{Cortile}, the same video is used to represent both the environment (1) and the related point of interest (Ingresso - 1.1). \figurename~\ref{fig:interest} shows some representative samples
of the $57$ points of interest acquired with HoloLens and GoPro Hero~4, whereas
\figurename~\ref{fig:negclass} reports some sample frames belonging to negative
locations and points of interest. As can be noted from the reported samples, the
GoPro device allow to acquire a larger amount of visual information, due to its
wide-angle field of view. On the contrary, data acquired using the HoloLens
device tends to exhibit more visual variability, due to the head-mounted point
of view, which suggests its better suitability for the recognition of objects of
interest and behavioral understanding.

\begin{table*}
	\centering
	\begin{tabular}{|c|c|lcc}
		\cline{1-2} \cline{4-5}
		\textbf{Environments} 		   & \textbf{Points of Interest} 		   &
		\multicolumn{1}{l|}{} & \multicolumn{1}{c|}{\textbf{Environments}}&
		\multicolumn{1}{c|}{\textbf{Points of Interest}} 		 \\ \cline{1-2} \cline{4-5} 
		Cortile (1)                            & Ingresso (1.1)                                    
		& \multicolumn{1}{l|}{} & \multicolumn{1}{l|}{}                         
		& \multicolumn{1}{c|}{PavimentoOriginale (6.3)}                                
		\\ \cline{1-2}
		\multirow{2}{*}{Scal. Monumentale (2)} & RampaS.Nicola (2.1)                  
		& \multicolumn{1}{l|}{} & \multicolumn{1}{c|}{}                         
		& \multicolumn{1}{c|}{PavimentoRestaurato (6.4)}                               
		\\
		& RampaS.Benedetto (2.2)                        & \multicolumn{1}{l|}{} &
		\multicolumn{1}{c|}{}                           &
		\multicolumn{1}{c|}{BassorilieviMancanti (6.5)}                               
		\\ \cline{1-2}
		& SimboloTreBiglie (3.1)                        & \multicolumn{1}{l|}{} &
		\multicolumn{1}{c|}{Aula S.Mazzarino (6)}       & \multicolumn{1}{c|}{LavamaniSx
			(6.6)}                                          \\
		& ChiostroLevante (3.2)                         & \multicolumn{1}{l|}{} &
		\multicolumn{1}{c|}{}                           & \multicolumn{1}{c|}{LavamaniDx
			(6.7)}                                          \\
		& Plastico (3.3)                                & \multicolumn{1}{l|}{} &
		\multicolumn{1}{c|}{}                           &
		\multicolumn{1}{c|}{TavoloRelatori (6.8)}                                     
		\\
		& Affresco (3.4)                                & \multicolumn{1}{l|}{} &
		\multicolumn{1}{c|}{}                           & \multicolumn{1}{c|}{Poltrone
			(6.9)}                                            \\ \cline{4-5} 
		& Finestra\_ChiostroLev. (3.5)                  & \multicolumn{1}{l|}{} &
		\multicolumn{1}{c|}{}                           & \multicolumn{1}{c|}{Edicola
			(7.1)}                                             \\
		Corridoi (3)                           & PortaCorodiNotte (3.6)               
		& \multicolumn{1}{l|}{} & \multicolumn{1}{c|}{}                         
		& \multicolumn{1}{c|}{PavimentoA (7.2)}                                        
		\\
		& TracciaPortone (3.7)                          & \multicolumn{1}{l|}{} &
		\multicolumn{1}{c|}{}                           & \multicolumn{1}{c|}{PavimentoB
			(7.3)}                                          \\
		& StanzaAbate (3.8)                             & \multicolumn{1}{l|}{} &
		\multicolumn{1}{c|}{Cucina (7)}                 &
		\multicolumn{1}{c|}{PassavivandePav.Orig. (7.4)}                              
		\\
		& CorridoioDiLevante (3.9)                      & \multicolumn{1}{l|}{} &
		\multicolumn{1}{c|}{}                           &
		\multicolumn{1}{c|}{AperturaPavimento (7.5)}                                  
		\\
		& CorridoioCorodiNotte (3.10)                   & \multicolumn{1}{l|}{} &
		\multicolumn{1}{c|}{}                           & \multicolumn{1}{c|}{Scala
			(7.6)}                                               \\
		& CorridoioOrologio (3.11)                      & \multicolumn{1}{l|}{} &
		\multicolumn{1}{c|}{}                           &
		\multicolumn{1}{c|}{SalaMetereologica (7.7)}                                  
		\\ \cline{1-2} \cline{4-5} 
		& Quadro (4.1)                                  & \multicolumn{1}{l|}{} &
		\multicolumn{1}{c|}{}                           & \multicolumn{1}{c|}{Doccione
			(8.1)}                                            \\
		Coro di Notte (4)                      & PavimentoOrig.Altare (4.2)           
		& \multicolumn{1}{l|}{} & \multicolumn{1}{c|}{}                         
		& \multicolumn{1}{c|}{VanoRaccoltaCenere (8.2)}                                
		\\
		& BalconeChiesa (4.3)                           & \multicolumn{1}{l|}{} &
		\multicolumn{1}{c|}{}                           & \multicolumn{1}{c|}{SalaRossa
			(8.3)}                                           \\ \cline{1-2}
		& PortaAulaS.Mazzarino (5.1)                    & \multicolumn{1}{l|}{} &
		\multicolumn{1}{c|}{}                           &
		\multicolumn{1}{c|}{ScalaCucina (8.4)}                                        
		\\
		& PortaIng.MuseoFabb. (5.2)                     & \multicolumn{1}{l|}{} &
		\multicolumn{1}{c|}{}                           &
		\multicolumn{1}{c|}{CucinaProvv. (8.5)}                                       
		\\
		& PortaAntirefettorio (5.3)                     & \multicolumn{1}{l|}{} &
		\multicolumn{1}{c|}{Ventre (8)}                 & \multicolumn{1}{c|}{Ghiacciaia
			(8.6)}                                          \\
		& PortaIngressoRef.Piccolo (5.4)                & \multicolumn{1}{l|}{} &
		\multicolumn{1}{c|}{}                           & \multicolumn{1}{c|}{Latrina
			(8.7)}                                             \\
		Antirefettorio (5)                     & Cupola (5.5)                         
		& \multicolumn{1}{l|}{} & \multicolumn{1}{c|}{}                         
		& \multicolumn{1}{c|}{OssaeScarti (8.8)}                                       
		\\
		& AperturaPavimento (5.6)                       & \multicolumn{1}{l|}{} &
		\multicolumn{1}{c|}{}                           & \multicolumn{1}{c|}{Pozzo
			(8.9)}                                               \\
		& S.Agata (5.7)                                 & \multicolumn{1}{l|}{} &
		\multicolumn{1}{c|}{}                           & \multicolumn{1}{c|}{Cisterna
			(8.10)}                                           \\
		& S.Scolastica (5.8)                            & \multicolumn{1}{l|}{} &
		\multicolumn{1}{c|}{}                           &
		\multicolumn{1}{c|}{BustoPietroTacchini (8.11)}                               
		\\ \cline{4-5} 
		& ArcoconFirma (5.9)                            & \multicolumn{1}{l|}{} &
		\multicolumn{1}{c|}{}                           &
		\multicolumn{1}{c|}{NicchiaePavimento (9.1)}                                  
		\\
		& BustoVaccarini (5.10)                         & \multicolumn{1}{l|}{} &
		\multicolumn{1}{c|}{Giardino Novizi (9)}        &
		\multicolumn{1}{c|}{TraccePalestra (9.2)}                                     
		\\ \cline{1-2}
		\multirow{2}{*}{Aula S.Mazzarino (6)}  & Quadro S.Mazzarino (6.1)             
		& \multicolumn{1}{l|}{} & \multicolumn{1}{c|}{}                         
		& \multicolumn{1}{c|}{PergolatoNovizi (9.3)}                                   
		\\ \cline{4-5} 
		& Affresco (6.2)                                &                       &     
		&                                    
		\\ \cline{1-2}
	\end{tabular}
	\caption{The table reports the list of all environments and the related points
		of interest contained. In parenthesis, we report the unique numerical code of
		the environment/point of interest.}
	\label{table:labels}
\end{table*}

The dataset is composed of separate training and test videos, which have been
collected following two different modalities. To collect the training set, we
acquired a set of videos (at least one) for each of the considered environments
and a set of videos for each of the considered points of interest.
Environment-related training videos have been acquired by an operator who had
been instructed to walk into the environment and look around to capture images
from different points of view. A similar procedure is employed to acquire
training videos for the different points of interest. In this case, the operator
has been instructed to walk around the object and look at it from different
points of view. For each camera, we collected a total of $12$ training videos
for the $9$ environments and $68$ videos for the $57$ points of interest. This
accounts to a total of $80$ training videos for each camera.
\tablename{~\ref{tab:training_videos}} summarizes the number of training videos
acquired for each environment.

\begin{table}
	\centering
	\begin{tabular}{|l|c|c|}
		\hline
		\textbf{Environment} & \multicolumn{1}{l|}{\textbf{\#Videos}} &
		\multicolumn{1}{l|}{\textbf{\#Frames}} \\ \hline
		1 Cortile              & 1                                          & 1171    
		\\
		2 Scalone Monumentale  & 6                                          & 13464   
		\\
		3 Corridoi             & 14                                         & 31037   
		\\
		4 Coro di Notte        & 7                                          & 12687   
		\\
		5 Antirefettorio       & 14                                         & 29918   
		\\
		6 Aula Santo Mazzarino & 10                                         & 31635   
		\\
		7 Cucina               & 10                                         & 31112   
		\\
		8 Ventre               & 14                                         & 68198   
		\\
		9 Giardino dei Novizi  & 4                                          & 10852   
		\\	\hline					
		\textbf{Total}        & 80				   						&
		230074									  \\					\hline
	\end{tabular}
	\caption{Training videos and total number of frames for each environment.}
	\label{tab:training_videos}
\end{table}

The test videos have been acquired by operators who have been asked to simulate
a visit to the cultural site. No specific information on where to move, what to
look, and for how long have been given to the operators. Test videos have been
acquired by three different subjects. Specifically, we acquired $7$ test video
per wearable camera, totaling $14$ test videos. Each HoloLens video is composed
by one or more video fragments due to the limit of recording time imposed by the
default  Hololens video capture application. 
Table~\ref{table:table_test} reports the list of test videos acquired using
HoloLens and GoPro devices. For each test video, we report the number of frames
and the list of environments the user visits during the acquisition of the
video.

\begin{table*}[]
	\centering
	\begin{tabular}{|l|c|l|llll}
		\cline{1-3} \cline{5-7}
		\multicolumn{3}{|c|}{\textbf{HoloLens}}                                       
		& \multicolumn{1}{l|}{} &
		\multicolumn{3}{c|}{\textbf{GoPro}}                                             
		\\ \cline{1-3} \cline{5-7} 
		\multicolumn{1}{|c|}{\textbf{Name}} & \textbf{\#Frames} &
		\multicolumn{1}{c|}{\textbf{Environments}}                  &
		\multicolumn{1}{l|}{} & \multicolumn{1}{c|}{\textbf{Name}} &
		\multicolumn{1}{c|}{\textbf{\#Frames}} &
		\multicolumn{1}{c|}{\textbf{Environments}}             \\ \hline
		Test1.0                             & 7202             &
		\multicolumn{1}{c|}{\multirow{2}{*}{1 - 2 - 3 - 4 - 5 - 6}} &
		\multicolumn{1}{l|}{} & \multicolumn{1}{l|}{Test1}         &
		\multicolumn{1}{c|}{14788}            & \multicolumn{1}{c|}{1 - 2 - 3 - 4}      
		\\ \cline{5-7} 
		Test1.1                             & 7202             & \multicolumn{1}{c|}{}
		& \multicolumn{1}{l|}{} &
		\multicolumn{1}{l|}{Test2}         & \multicolumn{1}{c|}{10503}            &
		\multicolumn{1}{c|}{1 - 2 - 3 - 5}                     \\ \cline{1-3}
		\cline{5-7} 
		Test2.0                             & 7203             & \multicolumn{1}{c|}{1
			- 2 - 3 - 4}                          & \multicolumn{1}{l|}{} &
		\multicolumn{1}{l|}{Test3}         & \multicolumn{1}{c|}{14491}            &
		\multicolumn{1}{c|}{1 - 2 - 3 - 5 - 9}                 \\ \cline{1-3}
		\cline{5-7} 
		Test3.0                             & 7202             &                      
		& \multicolumn{1}{l|}{} &
		\multicolumn{1}{l|}{Test4}         & \multicolumn{1}{c|}{36808}            &
		\multicolumn{1}{c|}{1 - 2 - 3 - 4 - 5 - 6 - 7 - 8 - 9} \\ \cline{5-7} 
		Test3.1                             & 7203             &                      
		& \multicolumn{1}{l|}{} &
		\multicolumn{1}{l|}{Test5}         & \multicolumn{1}{c|}{18788}            &
		\multicolumn{1}{c|}{1 - 2 - 3 - 5 - 7 - 8}             \\ \cline{5-7} 
		Test3.2                             & 7201             & \multicolumn{1}{c|}{1
			- 2 - 3 - 4 - 5 - 6 - 7 - 8 - 9}      & \multicolumn{1}{l|}{} &
		\multicolumn{1}{l|}{Test6}         & \multicolumn{1}{c|}{12661}            &
		\multicolumn{1}{c|}{2 - 3 - 4 - 8 - 9}                 \\ \cline{5-7} 
		Test3.3                             & 7202             &                      
		& \multicolumn{1}{l|}{} &
		\multicolumn{1}{l|}{Test7}         & \multicolumn{1}{c|}{38725}            &
		\multicolumn{1}{c|}{1 - 2 - 3 - 4 - 5 - 6 - 7 - 8 - 9} \\ \cline{5-7} 
		Test3.4                             & 5694             &                      
		& \multicolumn{1}{l|}{} &
		\multicolumn{1}{l|}{\textbf{Total}} & \multicolumn{1}{c|}{146764}            &
		\multicolumn{1}{c|}{} \\ \cline{5-7} 
		Test4.0                             & 7204             &                      
		&                       &                 
		&                                       &                     
		\\ \cline{1-3}
		Test4.1                             & 7202             &                      
		&                       &                 
		&                                       &                     
		\\
		Test4.2                             & 3281             & \multicolumn{1}{c|}{1
			- 2 - 3 - 4 - 5  - 7 - 8 - 9}         &                       &                 
		&                                       &                     
		\\
		Test4.3                             & 7202             &                      
		&                       &                 
		&                                       &                     
		\\
		Test4.4                             & 4845             &                      
		&                       &                 
		&                                       &                     
		\\ \cline{1-3}
		Test5.0                             & 6590             & \multirow{4}{*}{1 - 2
			- 3 - 4 - 5 - 6 - 7 - 8 - 9}          &                       &                 
		&                                       &                     
		\\
		Test5.1                             & 7202             &                      
		&                       &                 
		&                                       &                     
		\\
		Test5.2                             & 7202             &                      
		&                       &                 
		&                                       &                     
		\\
		Test5.3                             & 7201             &                      
		&                       &                 
		&                                       &                     
		\\ \cline{1-3}
		Test6.0                             & 7202             & \multicolumn{1}{c|}{1
			- 2 - 3}                              &                       &                 
		&                                       &                     
		\\ \cline{1-3}
		Test7.0                             & 7202             &
		\multicolumn{1}{c|}{\multirow{2}{*}{1 - 2 - 3 - 5}}         &                   
		&                                    &                                      
		&                                                        \\
		Test7.1                             & 2721             & \multicolumn{1}{c|}{}
		&                       &                 
		&                                       &                     
		\\ \cline{1-3}
		\textbf{Total}						& 131163		   & \multicolumn{1}{c|}{}                      
		&                       &                                    &  
		&                                           
		\\ \cline{1-3}
		
	\end{tabular}
	\caption{The list of test videos acquired using HoloLens (left) and GoPro
		(right). For each video, we report the number of frames and the list of
		environments visited by the user during the acquisition. The last rows report
		the total number of frames.}
	\label{table:table_test}
\end{table*}

The dataset, which we call UNICT-VEDI, is publicly available at our website:
\url{http://iplab.dmi.unict.it/VEDI/}. The reader is referred to the
supplementary material at the same page for
more details about the dataset.

\section{Egocentric Localization in a Cultural Site}
\label{sec:method}
We perform experiments and report baseline results related to the task of
localizing visitors from egocentric videos. To address the localization task, we
consider the approach proposed in ~\cite{furnari2018personal}.
This method is particularly suited for the considered task since it can be
trained with a small number of samples and includes a rejection option to
determine when the visitor is not located in any of the environments considered
at training time (i.e., when a frame belongs to the negative class). Moreover,
as detailed in~\cite{furnari2018personal}, the method achieves state of
the art results on the task of location-based temporal video segmentation, outperforming classic methods based on SVMs and local feature matching. At
test time, the algorithm segments an egocentric video into temporal segments
each associated to either one of the ``positive'' classes or, alternatively, the
``negative'' class. The approach is reviewed in the following section. The
reader is referred to~\cite{furnari2018personal} for more details.

\subsection{Method Review}
At training time, we define a set of $M$ ``positive'' classes, for which we
provide labeled training samples. In our case, this corresponds to the set of
training videos acquired for each of the considered environments. At test time,
an input egocentric video $\mathcal{V}=\{F_1,\ldots,F_N\}$ composed by $N$
frames $F_i$ is analyzed. Each frame of the video is assumed to belong to either
one of the considered $M$ positive classes or none of them. In the latter case,
the frame belongs to the ``negative class''. Since, negative training samples
are not assumed at training time, the algorithm has to detect which frames do not
belong to any of the positive classes and reject them. The goal of the system is
to divide the video into temporal segment, i.e., to produce a set of $P$ video
segments $\mathcal{S}=\{s_i\}_{1 \leq i\leq P}$, each associated with a class
(i.e., the room-level location). In particular, each segment is defined as
$s_i=\{s_i^s,s_i^e,s_i^c\}$, where $s_i^s$ represents the staring frame of the
segment, $s_i^e$ represents the ending frame of the segments and $s_i^c \in
\{0,\ldots,M\}$ represents the class of the segment ($s_i^c=0$ is the ``negative
class'', while $s_i^c=1,\ldots,M$ represent the ``positive'' classes).

\begin{figure*}
	\centering
	{\includegraphics[width=\linewidth]{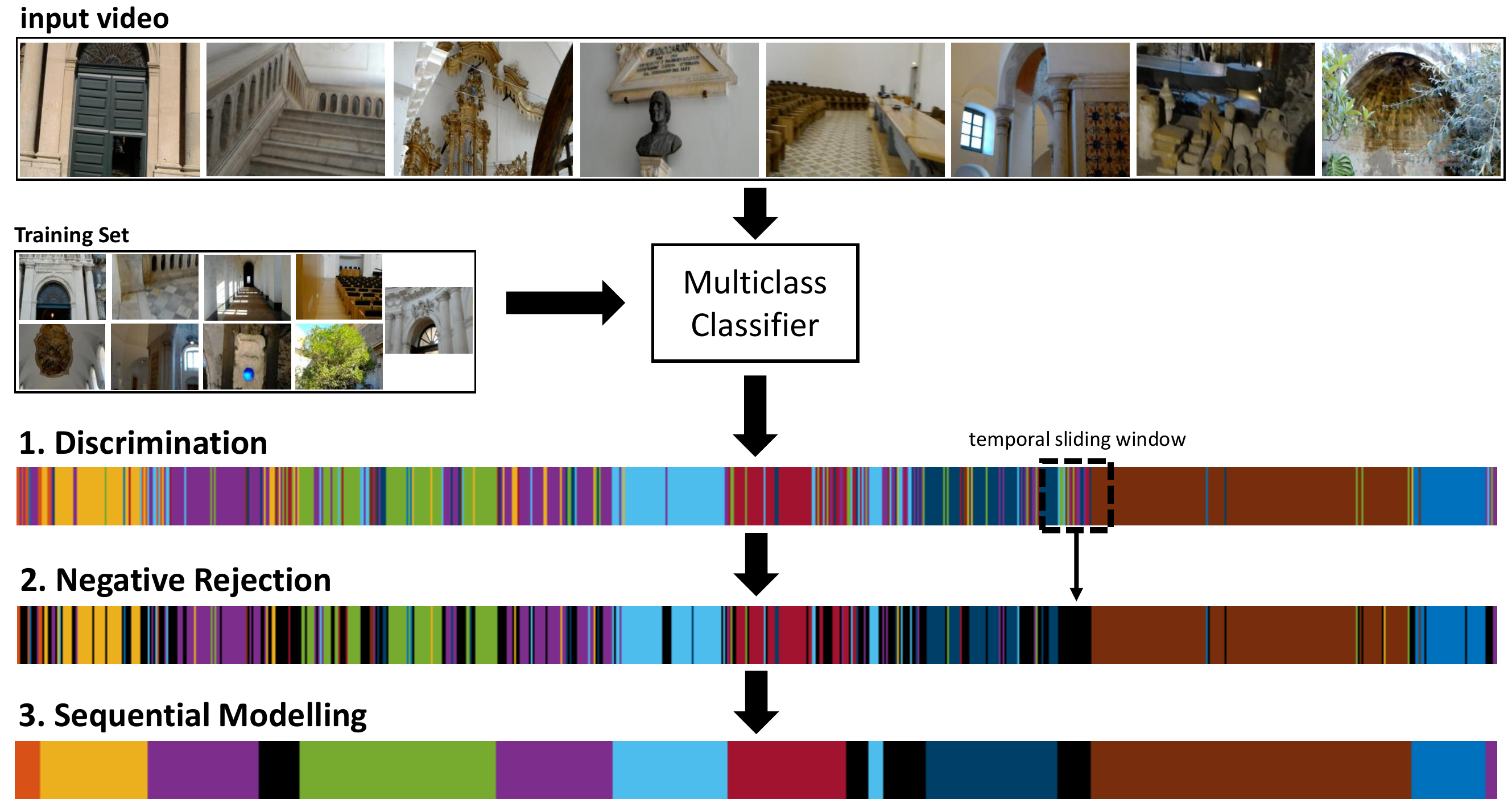}}	
	\qquad%
	\caption{Diagram of the considered room-based localization method consisting in
		three steps: 1) Discrimination, 2) Negative Rejection, 3) Sequential Modeling.}
	\label{fig:model}
\end{figure*}

The temporal segmentation of the input video is achieved in three steps:
discrimination, negative rejection and sequential modeling.
\figurename~\ref{fig:model} shows a diagram of the considered method, including
typical color-coded representations of the intermediate and final segmentation
output.

In the discrimination step, each frame of the video $F_i$ is assigned the most
probable class $y^*_i$ among the considered $M$ positive classes. In order to
perform such assignment, a multi-class classifier trained only on the positive samples is employed to estimate the
posterior probability distribution:
\begin{equation}
	\label{eq:posterior_multiclass}
	P(y_i|F_i, y_i \neq 0)
\end{equation}
where $y_i \neq 0$ indicates that the negative class is excluded from the
posterior probability. The most probable class $y^*_i$ is hence assigned using
the Maximum A Posteriori (MAP) criterion: $y^*_i = \arg\max_{y_i} P(y_i|I_F, y_i
\neq 0)$. Please note that, at this stage, the negative class is not considered.
The discrimination step allows to obtain a noisy assignment of labels to the
frames of the input video, as it is depicted in~\figurename~\ref{fig:model}.

The negative rejection step aims at identifying regions of the video in which
frames are likely to belong to the negative class. Since in an egocentric video
locations are deemed to change smoothly, regions containing negative frames are
likely to be characterized by noisy class assignments. This is expected since
the multi-class classifier used in the discrimination step had no knowledge of
the negative class. Moreover, consecutive frames of an egocentric video are
likely to contain uncorrelated visual content, due to fast head movements, which
would lead the multi-class classifier to pick a different class for each
negative frame. To leverage this consideration, the negative rejection step
quantifies the probability of each frame to belong to the negative class by
estimating the variation ratio (a measure of entropy) of the nominal
distribution of the assigned labels in a neighborhood of size $K$ centered at
the frame to be classified. Let
$\mathcal{Y}_{i}^K=\{y_{i-\lfloor\frac{K}{2}\rfloor},\ldots,
y_{i+\lfloor\frac{K}{2}\rfloor}\}$ be the set of positive labels assigned to the
frames comprised in a neighborhood of size $K$ centered at frame $F_i$. The
rejection probability for the frame $F_i$ is computed as the variation ratio of
the sample $\mathcal{Y}_{i}^K$:
\begin{eqnarray}
	\label{eq:negprob}
	P(y_i=0|F_i) =
	1-\frac{\sum_{k=i-\lfloor\frac{K}{2}\rfloor}^{i+\lfloor\frac{K}{2}\rfloor}{[y_k=mode(\mathcal{Y}_i^K)]}}{K}
\end{eqnarray}
where $[\cdot]$ is the Iverson bracket, $y_k \in \mathcal{Y}_i^K$ and
$mode(\mathcal{Y}_i^K)$ is the most frequent label of $\mathcal{Y}_i^K$. Since
$y_i=0$ and $y_i\neq 0$ are disjoint events, the posterior probability defined
in Equation~\eqref{eq:posterior_multiclass} can be easily merged to the
probability defined in Equation~\eqref{eq:negprob} to estimate the posterior
probability $P(y_i|F_i)$. Note that this is a posterior probability over the $M$
positive classes, plus the negative one. The MAP criterion can be used to assign
each frame $F_i$ the most probable class $y_i$ using the posterior probability
$P(y_i|F_i)$ (see ~\figurename~\ref{fig:model}). Please note that, in
this case, the assigned labels include the negative class.

The label assignment obtained in the negative rejection step is still a noisy
one (see~\figurename~\ref{fig:model}). The sequential modeling step smooths the
segmentation result enforcing temporal coherence among neighboring predictions.
This is done employing a Hidden Markov Model (HMM)~\cite{bishop2006pattern} with
$M+1$ states ($M$ ``positive'' classes plus the ``negative'' one). The HMM
models the conditional probability of the labels
$\mathcal{L}=\{y_1,\ldots,y_N\}$ given the video $\mathcal{V}$:
\begin{equation}
	\label{eq:hmm}
	P(\mathcal{L}|\mathcal{V}) \propto \prod_{i=2}^N P(y_i|y_{i-1}) \prod_{i=1}^N
	P(y_i|F_i)
\end{equation}
where $P(y_i|I_i)$ models the emission probability (i.e., the probability of
being in state $y_i$ given the frame $F_i$). The state transition
probabilities $P(y_i|y_{i-1})$ are modeled defining an ``almost identity
matrix'' which encourages the model to rarely allow for state changes:
\begin{eqnarray}
	\label{eq:transition}
	P(y_i|y_{i-1}) = \begin{cases}
		\varepsilon, & \mbox{if } y_i \neq y_{i-1} \\
		1-M\varepsilon, & \mbox{otherwise}
	\end{cases}
\end{eqnarray}
The definition above depends on a parameter $\varepsilon$ which controls the
amount of smoothing in the predictions. The optimal set of labels $\mathcal{L}$
according to the defined HMM can be obtained using the Viterbi algorithm (see~\figurename~\ref{fig:model} - 3. Sequential Modeling). The segmentation $\mathcal{S}$ is
finally obtained by considering the connected components of the optimal set of
labels $\mathcal{L}$.

\section{Experimental Settings}
\label{sec:experiments}
In this section, we discuss the experimental settings used for the experiments.
To assess the potential of the two
considered devices, we perform experiments separately on data acquired using
HoloLens and GoPro by training and testing two separate models.

To setup the method reviewed in Section~\ref{sec:method}, it is necessary to
train a multi-class classifier to discriminate between the $M$ positive classes (i.e., environments).
We implement this component fine-tuning a VGG19 Convolutional Neural Network
(CNN) pre-trained on the ImageNet dataset~\cite{Simonyan14c} to discriminate
between the $9$ considered classes (\textit{Cortile, Scalone Monumentale, Corridoi, Coro
di Notte, Antirefettorio, Aula Santo Mazzarino, Cucina, Ventre} e \textit{Giardino dei
Novizi}). To compose our training set, we first considered all image frames
belonging to the training videos collected for each environment
(Figure~\ref{tab:training_videos}). We augment the frames of each of the
considered environments including also frames from the training videos collected
for the points of interest contained in the environment. We finally select
exactly $10000$ frames for each class, except for the ``Cortile'' (1) class
which contained only $1171$ frames (in this case all frames have been
considered). As previously discussed, the same video is used for both the environment ``Cortile'' (1) and  
point of interest \textit{Ingresso (1.1)}. Therefore, it was not possible to gather more frames from videos related to the points of interest. To validate the performances of the classifier, we randomly select $30\%$ of the
training samples to obtain a validation set. Please note that the CNN classifier
is trained solely on positive data and no negatives are employed at this stage.
To select the optimal values for the parameters $K$ (neighborhood size for
negative rejection) and $\varepsilon$ (HMM smoothing parameter), we carry out a
grid search on one of the test videos, which is used as ``validation video''.
Specifically, we consider $K \in \{50,100,300\}$ and $\varepsilon \in
[10^{-300},10^{-2}]$ for the grid search. We select ``Test 3'' as a validation video for the algorithms trained on data acquired with HoloLens and ``Test 4'' for the experiments related to data acquired using the GoPro camera. These two videos are selected since they contain all the classes and, overall, similar content (see Table~\ref{table:table_test}). Moreover, the two videos have been acquired simultaneously by the same operator to provide similar material for validation. The grid search led to the
selection of the following parameter values: $K=50; \varepsilon= 10^{-152}$ 
in the case of the experiments performed on HoloLens data and 
$K=300; \varepsilon= 10^{-171}$ for the experiments performed on the GoPro data.
We used the Caffe framework~\cite{jia2014caffe} to train the CNN models. All the data, code and trained models useful to replicate the work are
publicly available for download at our web page
\url{http://iplab.dmi.unict.it/VEDI/}.

All experiments have been evaluated according to two complementary measures:
$FF_1$ and $ASF_1$~\cite{furnari2018personal}. The $FF_1$ is a frame-based measure obtained computing the
frame-wise $F_1$ score for each class. This measure essentially assesses how many
frames have been correctly classified without taking into account the temporal
structure of the prediction. A high $FF_1$ score indicates that the method is able to estimate the number of frames belonging to a given class in a video. This is useful to assess, for instance, how much times has been spent at a given location, regardless the temporal structure. To assess how well the algorithm can split the
input videos into coherent segments, we also use the $ASF_1$ score, which
measures how accurate the output segmentation is with respect to ground truth.
$FF_1$ and $ASF_1$ scores are computed per class. We also report overall $mFF_1$
and $mASF_1$ scores obtained by averaging class-related scores. The reader is
referred to~\cite{furnari2018personal} for details on the implementation of such
scores. An implementation of these measures is available at the following link: \url{http://iplab.dmi.unict.it/PersonalLocationSegmentation/}.

\begin{table}[]
	\centering
	\begin{adjustbox}{width=\linewidth}
	\begin{tabular}{lccccccc}
		\hline
		\multicolumn{1}{l|}{Class}                  & Test1 & Test2 & Test4 & Test5 & Test6 & \multicolumn{1}{c|}{Test7} & \multicolumn{1}{l}{AVG}   \\ \hline
		\multicolumn{1}{l|}{1 Cortile}              & 0.94  & 0.00     & 0.93  & 0.00     & 0.95  & \multicolumn{1}{c|}{0.77}  & 0.59                     \\
		\multicolumn{1}{l|}{2 Scalone Monumentale}  & 0.99  & 0.98  & 0.99  & 0.95  & 0.98  & \multicolumn{1}{c|}{0.85}  & 0.96                     \\
		\multicolumn{1}{l|}{3 Corridoi}             & 0.93  & 0.93  & 0.95  & 0.85  & 0.99  & \multicolumn{1}{c|}{0.85}  & 0.92                     \\
		\multicolumn{1}{l|}{4 Coro Di Notte}        & 0.94  & 0.94  & 0.89  & 0.87  & /     & \multicolumn{1}{c|}{/}     & 0.91                     \\
		\multicolumn{1}{l|}{5 Antirefettorio}       & 0.94  & /     & 0.96  & 0.94  & /     & \multicolumn{1}{c|}{0.94}  & 0.95                     \\
		\multicolumn{1}{l|}{6 Aula Santo Mazzarino} & 0.99  & /     & /     & 0.98  & /     & \multicolumn{1}{c|}{/}     & 0.99                     \\
		\multicolumn{1}{l|}{7 Cucina}               & /     & /     & 0.65  & 0.75  & /     & \multicolumn{1}{c|}{/}     & 0.70                     \\
		\multicolumn{1}{l|}{8 Ventre}               & /     & /     & 0.92  & 0.99  & /     & \multicolumn{1}{c|}{/}     & 0.96                     \\
		\multicolumn{1}{l|}{9 Giardino dei Novizi}  & /     & /     & 0.95  & 0.71  & /     & \multicolumn{1}{c|}{/}     & 0.83                     \\
		\multicolumn{1}{l|}{Negatives}              & 0.56  & 0.50   & 0.26  & 0.37  & /     & \multicolumn{1}{c|}{/}     & \multicolumn{1}{c}{0.42} \\ \hline
		\multicolumn{1}{c|}{$mFF_1$}                   & 0.90 & 0.67 & 0.83 & 0.74 & 0.97 & \multicolumn{1}{c|}{0.85}                      & 0.82                    
	\end{tabular}
\end{adjustbox}
	\caption{Frame-based $FF_1$ scores of the considered method on data acquired using the HoloLens device. The ``/'' sign indicates that no samples from that class were present in the test video.}
	\label{tab:mff1_HoloLens}
\end{table}

\begin{table}[]
	\centering
	\begin{adjustbox}{width=\linewidth}
	\begin{tabular}{lccccccc}\hline
		\multicolumn{1}{l|}{Class}                  & Test1 & Test2 & Test4 & Test5 & Test6 & \multicolumn{1}{c|}{Test7} & \multicolumn{1}{l}{AVG} \\ \hline
		\multicolumn{1}{l|}{1 Cortile}              & 0.89  & 0.00     & 0.86  & 0.00     & 0.90   & \multicolumn{1}{c|}{0.62}  & 0.48                    \\
		\multicolumn{1}{l|}{2 Scalone Monumentale}  & 0.97  & 0.95  & 0.97  & 0.86  & 0.97  & \multicolumn{1}{c|}{0.74}  & 0.90                    \\
		\multicolumn{1}{l|}{3 Corridoi}             & 0.85  & 0.87  & 0.84  & 0.69  & 0.99  & \multicolumn{1}{c|}{0.58}  & 0.79                    \\
		\multicolumn{1}{l|}{4 Coro Di Notte}        & 0.81  & 0.90   & 0.78  & 0.31  & /     & \multicolumn{1}{c|}{/}     & 0.66                    \\
		\multicolumn{1}{l|}{5 Antirefettorio}       & 0.88  & /     & 0.93  & 0.66  & /     & \multicolumn{1}{c|}{0.90}   & 0.83                    \\
		\multicolumn{1}{l|}{6 Aula Santo Mazzarino} & 0.99  & /     & /     & 0.96  & /     & \multicolumn{1}{c|}{/}     & 0.96                    \\
		\multicolumn{1}{l|}{7 Cucina}               & /     & /     & 0.48  & 0.57  & /     & \multicolumn{1}{c|}{/}     & 0.53                    \\
		\multicolumn{1}{l|}{8 Ventre}               & /     & /     & 0.85  & 0.99  & /     & \multicolumn{1}{c|}{/}     & 0.92                    \\
		\multicolumn{1}{l|}{9 Giardino dei Novizi}  & /     & /     & 0.90  & 0.66  & /     & \multicolumn{1}{c|}{/}     & 0.78                    \\
		\multicolumn{1}{l|}{Negatives}              & 0.50   & 0.39  & 0.12  & 0.3   & /     & \multicolumn{1}{c|}{/}     & 0.27                    \\ \hline
		\multicolumn{1}{c|}{$mASF_1$}                  & 0.84  & 0.62  & 0.71  & 0.60  & 0.95  & \multicolumn{1}{c|}{0.71}                       & 0.71                   
	\end{tabular}
\end{adjustbox}
	\caption{Segment-based $ASF_1$ scores of the considered method on data acquired using the HoloLens device. The ``/'' sign indicates that no samples from that class were present in the test video.}
	\label{tab:masf_HoloLens}
\end{table}

\section{Results}
\label{sec:results}

\tablename~\ref{tab:mff1_HoloLens} and \tablename~\ref{tab:masf_HoloLens} report
the $mFF_1$ and $mASF_1$ scores obtained using data acquired using the HoloLens device.
Please note that all algorithms have been trained using only training
data acquired with the HoloLens device (no GoPro data has been used). ``Test 3''
is excluded from the table since it has been used for validation. The tables
report the $FF_1$ and $ASF_1$ scores for each class and each test video, average
per-class $FF_1$ and $ASF_1$ scores across videos, overall $mFF_1$ and $mASF_1$
scores for each test video and the average $mFF_1$ and $mASF_1$ scores which
summarize the performances over the whole test set. As can be noted from
both tables, some environments such as ``Cortile'', ``Cucina'' and ``Giardino
dei Novizi'' are harder to recognize than others. This is due to the greater
variability characterizing such environments. In particular, ``Cortile'' and
``Giardino dei Novizi'' are outdoor environments, while all the others are indoor
environments. It should be noted that, as discussed before, the two considered measures ($mFF_1$ and $mASF_1$) capture different abilities of the algorithm. For instance, some environments (e.g., ``Corridoi'' and ``Coro di Notte'') report high $mFF_1$, and lower $mASF_1$. This indicates that the method is able to quantify the overall amount of time spent at the considered location, but temporal structure of the segments is not correctly retrieved.
The average $mASF_1$ of $0.71$ and $mFF_1$ of $0.83$ obtained over the whole test set indicate that the proposed approach can be already useful to provide localization information to the visitor or for later analysis, e.g., to estimate how much time has been spent by a visitor at a given location, how many times a given environment has been visited, or what are the paths preferred by visitors.

\begin{figure}[t]
	\centering
	\includegraphics[width=\linewidth]{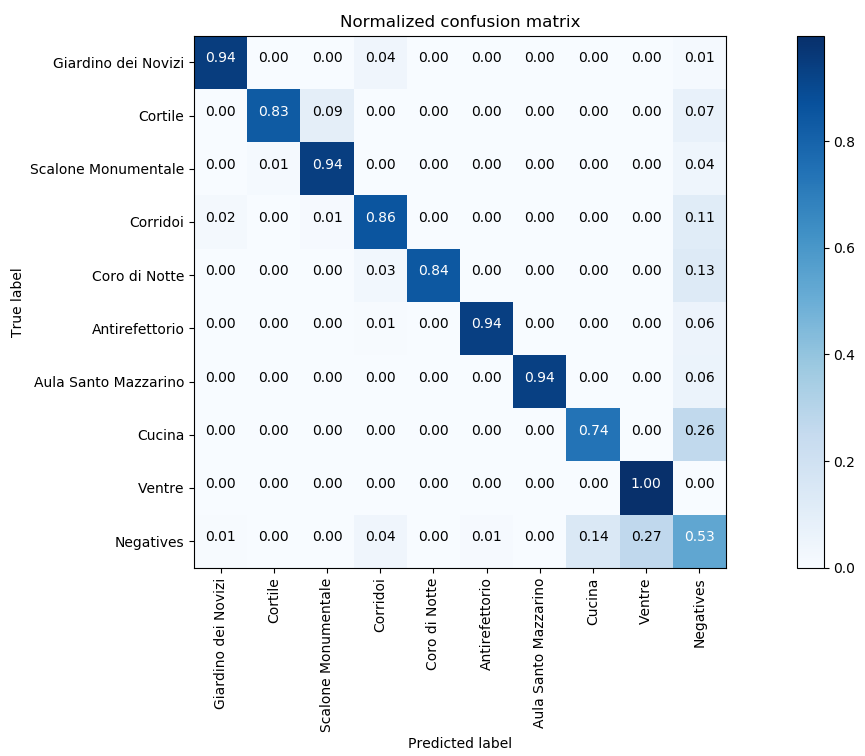}
	\caption{Confusion matrix of the considered method trained and tested on the HoloLens data.}
	\label{fig:cf_HoloLens}
\end{figure}

\figurename~\ref{fig:cf_HoloLens} reports the confusion matrix of the system on the HoloLens test set. The confusion matrix does not include frames from the ``Test 3'' video, which has been used for validation. The matrix confirms how some distinctive environments are well recognized, while others are more challenging. The matrix also suggests that most of the error is due to the challenging rejection of negative samples. Other minor source of errors are the ``Giardino dei Novizi - Corridoi'', ``Cortile - Scalone Monumentale'' and ``Coro di Notte - Corridoi'' class pairs. We note that the considered pairs are neighboring locations, which suggests that the error is due to small inaccuracies in the temporal segmentation.

We also report experiments performed on the GoPro data. To compare the effects of using different acquisition devices and wearing modalities, we replicate the same pipeline used for the experiments reported in the previous section. Hence, we trained and tested the same algorithms on data acquired using the GoPro device.

\tablename~\ref{tab:ff_gopro} and \tablename~\ref{tab:asf_gopro} report the $mFF_1$ and $mASF_1$ scores for the test videos acquired using the GoPro device. Results related to the ``Test 4'' validation video are excluded from the tables. The method allows to obtain overall similar performances for the different devices (an average $mFF_1$ score of $0.81$ in the case of GoPro, vs $0.82$ in the case of HoloLens and an average $mASF_1$ score of $0.71$ vs $0.71$). However, $mFF1$ performances on the single test videos are distributed differently (e.g., ``Test 1'' has a $mFF_1$ score of $0.90$ in the case of HoloLens data and a $mFF_1$ score of $0.67$ in the case of GoPro data).

\begin{table}[]
	\centering
	\begin{adjustbox}{width=\linewidth}
	\begin{tabular}{lccccccc}\hline
		\multicolumn{1}{l|}{Class}                 & Test1                    & Test2                 & Test3                 & Test5                    & Test6                 & \multicolumn{1}{c|}{Test7} & \multicolumn{1}{c}{AVG}   \\ \hline
		\multicolumn{1}{l|}{1 Cortile}              & 0.00                        & 0.97                  & 0.95                  & 0.92                     & /                     & \multicolumn{1}{c|}{0.00}     & 0.57                     \\
		\multicolumn{1}{l|}{2 Scalone Monumentale}  & 0.92                     & 0.92                  & 0.99                  & 0.99                     & 0.96                  & \multicolumn{1}{c|}{0.90}   & 0.95                     \\
		\multicolumn{1}{l|}{3 Corridoi}             & 0.90                      & 0.97                  & 0.99                  & 0.99                     & 0.97                  & \multicolumn{1}{c|}{0.98}  & 0.97                     \\
		\multicolumn{1}{l|}{4 Coro Di Notte}        & 0.89                     & /                     & /                     & /                        & 0.98                  & \multicolumn{1}{c|}{0.88}  & 0.92                     \\
		\multicolumn{1}{l|}{5 Antirefettorio}       & /                        & 0.99                  & 0.98                  & 0.96                     & /                     & \multicolumn{1}{c|}{0.87}  & 0.95                     \\
		\multicolumn{1}{l|}{6 Aula Santo Mazzarino} & /                        & /                     & /                     & /                        & /                     & \multicolumn{1}{c|}{0.90}   & 0.90                     \\
		\multicolumn{1}{l|}{7 Cucina}               & /                        & /                     & /                     & 0.89                     & /                     & \multicolumn{1}{c|}{0.83}  & 0.86                     \\
		\multicolumn{1}{l|}{8 Ventre}               & /                        & /                     & /                     & 0.99                     & 0.67                  & \multicolumn{1}{c|}{0.97}  & 0.88                     \\
		\multicolumn{1}{l|}{9 Giardino dei Novizi}  & /                        & /                     & 0.99                  & /                        & 0.95                  & \multicolumn{1}{c|}{0.52}  & 0.82                     \\
		\multicolumn{1}{l|}{Negatives}              & \multicolumn{1}{c}{0.47} & \multicolumn{1}{c}{/} & \multicolumn{1}{c}{/} & \multicolumn{1}{c}{0.52} & \multicolumn{1}{c}{0.00} & \multicolumn{1}{c|}{0.21}  & \multicolumn{1}{c}{0.30} \\ \hline
		\multicolumn{1}{c|}{$mFF_1$}                   & 0.67                    & 0.96                 & 0.98                 & 0.90                    & 0.76                 & \multicolumn{1}{c|}{0.71}                      & 0.81                    
	\end{tabular}
\end{adjustbox}
	\caption{Frame-based $FF_1$ scores of the considered method on data acquired using the GoPro device. The ``/'' sign indicates that no samples from that class were present in the test video.}
	\label{tab:ff_gopro}
\end{table}

\begin{table}[]
	\centering
	\begin{adjustbox}{width=\linewidth}
	\begin{tabular}{lccccccc}\hline
		\multicolumn{1}{l|}{Class}                 & Test1                    & Test2                 & Test3                 & Test5                    & Test6                 & \multicolumn{1}{c|}{Test7} & \multicolumn{1}{l}{AVG}  \\ \hline
		\multicolumn{1}{l|}{1 Cortile}              & 0.00                        & 0.94                  & 0.91                  & 0.85                     & /                     & \multicolumn{1}{c|}{0}     & 0.68                     \\
		\multicolumn{1}{l|}{2 Scalone Monumentale}  & 0.85                     & 0.65                  & 0.98                  & 0.97                     & 0.92                  & \multicolumn{1}{c|}{0.73}  & 0.85                     \\
		\multicolumn{1}{l|}{3 Corridoi}             & 0.86                     & 0.60                   & 0.97                  & 0.99                     & 0.92                  & \multicolumn{1}{c|}{0.93}  & 0.88                     \\
		\multicolumn{1}{l|}{4 Coro Di Notte}        & 0.76                     & /                     & /                     & /                        & 0.96                  & \multicolumn{1}{c|}{0.2}   & 0.58                     \\
		\multicolumn{1}{l|}{5 Antirefettorio}       & /                        & 0.97                  & 0.96                  & 0.92                     & /                     & \multicolumn{1}{c|}{0.66}  & 0.88                     \\
		\multicolumn{1}{l|}{6 Aula Santo Mazzarino} & /                        & /                     & /                     & /                        & /                     & \multicolumn{1}{c|}{0.81}  & 0.81                     \\
		\multicolumn{1}{l|}{7 Cucina}               & /                        & /                     & /                     & 0.79                     & /                     & \multicolumn{1}{c|}{0.68}  & 0.74                     \\
		\multicolumn{1}{l|}{8 Ventre}               & /                        & /                     & /                     & 0.99                     & 0.5                   & \multicolumn{1}{c|}{0.48}  & 0.66                     \\
		\multicolumn{1}{l|}{9 Giardino dei Novizi}  & /                        & /                     & 0.97                  & /                        & 0.91                  & \multicolumn{1}{c|}{0.61}  & 0.83                     \\
		\multicolumn{1}{l|}{Negatives}              & \multicolumn{1}{c}{0.48} & \multicolumn{1}{c}{/} & \multicolumn{1}{c}{/} & \multicolumn{1}{c}{0.45} & \multicolumn{1}{c}{0.00} & \multicolumn{1}{c|}{0.24}  & \multicolumn{1}{c}{0.23} \\ \hline
		\multicolumn{1}{c|}{$mASF_1$}                  & 0.59                     & 0.79                  & 0.96                  & 0.85                     & 0.70                  & \multicolumn{1}{c|}{0.53}                       & 0.71                    
	\end{tabular}
\end{adjustbox}
	\caption{Segment-based $ASF_1$ scores of the considered method on data acquired using the GoPro device. The ``/'' sign indicates that no samples from that class were present in the test video.}
	\label{tab:asf_gopro}
\end{table}

\begin{figure}[t]
	\centering
	\includegraphics[width=\linewidth]{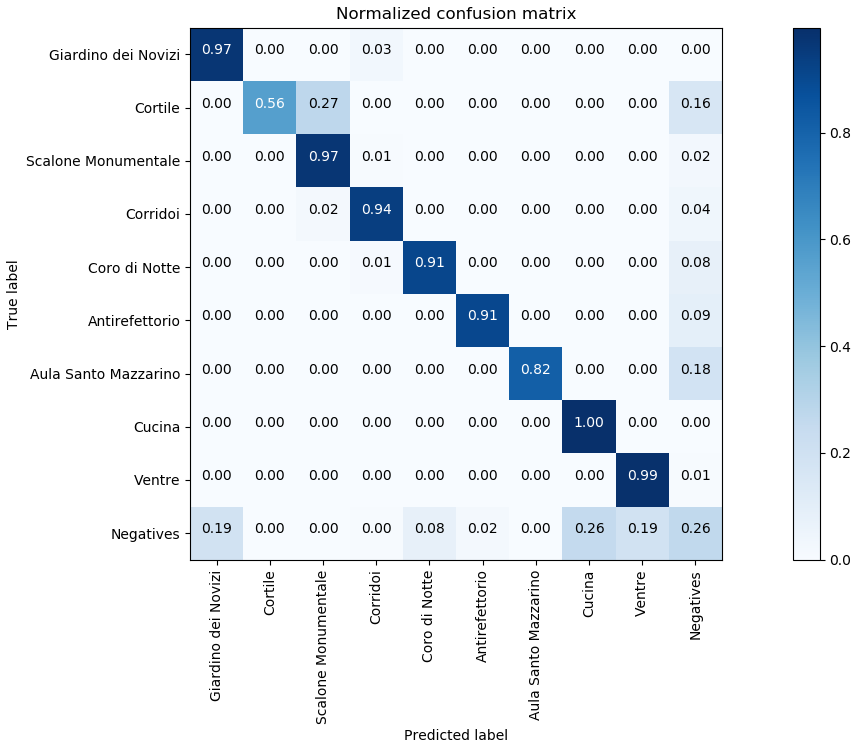}
	\caption{Confusion matrix of the results of the considered method on the GoPro test set.}
	\label{fig:cf_gopro}
\end{figure}

\figurename~\ref{fig:cf_gopro} reports the confusion matrix of the method over the GoPro test set, excluding the ``Test 4'' video (used for validation). Also in this case, errors are distributed differently with respect to the case of HoloLens data. In particular, the confusion between ``Cortile'' and ``Scalone Monumentale'' is much larger than in the case of HoloLens data, while other classes such as ``Cucina'' report better performance on the GoPro data. Moreover, the rejection of negatives is much worse performing in the case of GoPro data. These differences are due to the different way the GoPro camera captures the visual data. On the one hand, GoPro is characterized by a larger field of view, which allows to gather supplementary information for location recognition. On the other hand, the dynamic field of view of the head-mounted HoloLens device, allows to capture diverse and distinctive elements of the environment and allows for better rejection of negative frames increasing the amount of discrimination entropy in unknown environments.

\tablename~\ref{tab:ff_vs} and \tablename~\ref{tab:asf_vs} summarize and compare the results obtained training the algorithm on the two sets of data. Specifically, the tables report the average $FF_1$ and $ASF_1$ scores obtained in the three steps of the algorithm. As can be noted, significantly better discrimination is overall obtained using GoPro data ($0.88 mFF_1$ vs $0.73 mFF_1$). This is probably due to the wider Field Of View of the GoPro camera, which allows to capture more information about the surrounding environment (see~\figurename~\ref{fig:environments}). Rejecting negative frames is a challenging task, which leads to degraded performances both in the case of frame-based measures (\tablename~\ref{tab:ff_vs}) and segment-based ones (\tablename~\ref{tab:asf_vs}). Interestingly, the negative rejection step works best on HoloLens data ($0.66 mFF_1$ vs $0.54 mFF_1$, and $0.24 FF_1$ vs $0.18 FF_1$ for the negative class). This result confirms the aforementioned observation that HoloLens data allows to acquire more distinctive details about the scene, thus allowing for more entropy when in the presence of unknown environments. The sequential modeling step, finally balances out the results, allowing HoloLens and GoPro to achieve similar performances. 

\begin{table}[]
	\centering

	\begin{adjustbox}{width=\linewidth}
		\begin{tabular}{lcccccc}\hline
			\multicolumn{1}{l|}{}                       & \multicolumn{2}{c|}{Discrimination}           & \multicolumn{2}{c|}{Rejection}        & \multicolumn{2}{c}{Seq. Modeling} \\ \hline
			\multicolumn{1}{l|}{Class}                  & HoloLens & \multicolumn{1}{c|}{GoPro}         & HoloLens & \multicolumn{1}{c|}{GoPro} & HoloLens      & GoPro             \\ \hline
			\multicolumn{1}{l|}{1 Cortile}              & 0.50     & \multicolumn{1}{c|}{0.84}          & 0.45     & \multicolumn{1}{c|}{0.25}  & 0.59          & 0.57              \\
			\multicolumn{1}{l|}{2 Scalone Monumentale}  & 0.81     & \multicolumn{1}{c|}{0.93}          & 0.84     & \multicolumn{1}{c|}{0.91}  & 0.96          & 0.95              \\
			\multicolumn{1}{l|}{3 Corridoi}             & 0.77     & \multicolumn{1}{c|}{0.92}          & 0.69     & \multicolumn{1}{c|}{0.83}  & 0.92          & 0.97              \\
			\multicolumn{1}{l|}{4 Coro Di Notte}        & 0.71     & \multicolumn{1}{c|}{0.91}          & 0.67     & \multicolumn{1}{c|}{0.64}  & 0.91          & 0.92              \\
			\multicolumn{1}{l|}{5 Antirefettorio}       & 0.66     & \multicolumn{1}{c|}{0.83}          & 0.73     & \multicolumn{1}{c|}{0.62}  & 0.95          & 0.95              \\
			\multicolumn{1}{l|}{6 Aula Santo Mazzarino} & 0.69     & \multicolumn{1}{c|}{0.81}          & 0.65     & \multicolumn{1}{c|}{0.23}  & 0.99          & 0.90              \\
			\multicolumn{1}{l|}{7 Cucina}               & 0.72     & \multicolumn{1}{c|}{0.90}          & 0.60     & \multicolumn{1}{c|}{0.11}  & 0.70          & 0.86              \\
			\multicolumn{1}{l|}{8 Ventre}               & 0.97     & \multicolumn{1}{c|}{0.99}          & 0.94     & \multicolumn{1}{c|}{0.86}  & 0.96          & 0.88              \\
			\multicolumn{1}{l|}{9 Giardino dei Novizi}  & 0.79     & \multicolumn{1}{c|}{0.82}          & 0.79     & \multicolumn{1}{c|}{0.78}  & 0.83          & 0.82              \\
			\multicolumn{1}{l|}{Negatives}                    & /        & \multicolumn{1}{c|}{/}             & 0.24     & \multicolumn{1}{c|}{0.18}  & 0.42          & 0.30              \\ \hline
			\multicolumn{1}{c|}{$mFF_1$}                   & 0.73     & \multicolumn{1}{c|}{0.88} & 0.66     & \multicolumn{1}{c|}{0.54}  & 0.82          & 0.81    
		\end{tabular}%
	\end{adjustbox}

	\caption{Comparative table of average $FF_1$ scores for the considered method trained and tested on HoloLens and GoPro data. The table reports scores for the overall method (seq. modeling column), as well as for the two intermediate steps of Discrimination and Rejection.}
	\label{tab:ff_vs}
\end{table}

\begin{table}[]
	\centering
	\begin{adjustbox}{width=\linewidth}

		\begin{tabular}{lcccccc}\hline
			\multicolumn{1}{l|}{}                       & \multicolumn{2}{c|}{Discrimination}   & \multicolumn{2}{c|}{Rejection}        & \multicolumn{2}{c}{Seq. Modeling} \\ \hline
			\multicolumn{1}{l|}{Class}                  & HoloLens & \multicolumn{1}{c|}{GoPro} & HoloLens & \multicolumn{1}{c|}{GoPro} & HoloLens        & GoPro           \\ \hline
			\multicolumn{1}{l|}{1 Cortile}              & 0.01    & \multicolumn{1}{c|}{0.15} & 0.02    & \multicolumn{1}{c|}{0.03} & 0.48           & 0.68           \\
			\multicolumn{1}{l|}{2 Scalone Monumentale}  & 0.01    & \multicolumn{1}{c|}{0.05} & 0.01    & \multicolumn{1}{c|}{0.03} & 0.90           & 0.85           \\
			\multicolumn{1}{l|}{3 Corridoi}             & 0.00    & \multicolumn{1}{c|}{0.02} & 0.00    & \multicolumn{1}{c|}{0.01} & 0.79           & 0.88           \\
			\multicolumn{1}{l|}{4 Coro Di Notte}        & 0.00    & \multicolumn{1}{c|}{0.00} & 0.01    & \multicolumn{1}{c|}{0.01} & 0.66           & 0.58           \\
			\multicolumn{1}{l|}{5 Antirefettorio}       & 0.00    & \multicolumn{1}{c|}{0.02} & 0.01    & \multicolumn{1}{c|}{0.01} & 0.83           & 0.88           \\
			\multicolumn{1}{l|}{6 Aula Santo Mazzarino} & 0.00    & \multicolumn{1}{c|}{0.00} & 0.00    & \multicolumn{1}{c|}{0.00} & 0.96           & 0.81           \\
			\multicolumn{1}{l|}{7 Cucina}               & 0.00    & \multicolumn{1}{c|}{0.01} & 0.00    & \multicolumn{1}{c|}{0.00} & 0.52           & 0.74           \\
			\multicolumn{1}{l|}{8 Ventre}               & 0.00    & \multicolumn{1}{c|}{0.32} & 0.00    & \multicolumn{1}{c|}{0.03} & 0.92           & 0.66           \\
			\multicolumn{1}{l|}{9 Giardino dei Novizi}  & 0.01    & \multicolumn{1}{c|}{0.15} & 0.01    & \multicolumn{1}{c|}{0.04} & 0.78           & 0.83           \\
			\multicolumn{1}{l|}{Negatives}                    & /        & \multicolumn{1}{c|}{/}     & 0.01    & \multicolumn{1}{c|}{0.00} & 0.27           & 0.23           \\ \hline
			\multicolumn{1}{c|}{$mASF_1$}                  & 0.00    & \multicolumn{1}{c|}{0.08} & 0.01    & \multicolumn{1}{c|}{0.02} & 0.71  & 0.71
		\end{tabular}%
	\end{adjustbox}

	\caption{Comparative table of average $AFF_1$ scores for the considered method trained and tested on HoloLens and GoPro data. The table reports scores for the overall method (seq. modeling column), as well as for the two intermediate steps of Discrimination and Rejection.}
	\label{tab:asf_vs}
\end{table}

\begin{figure}[t]
	\centering
	\includegraphics[width=\linewidth]{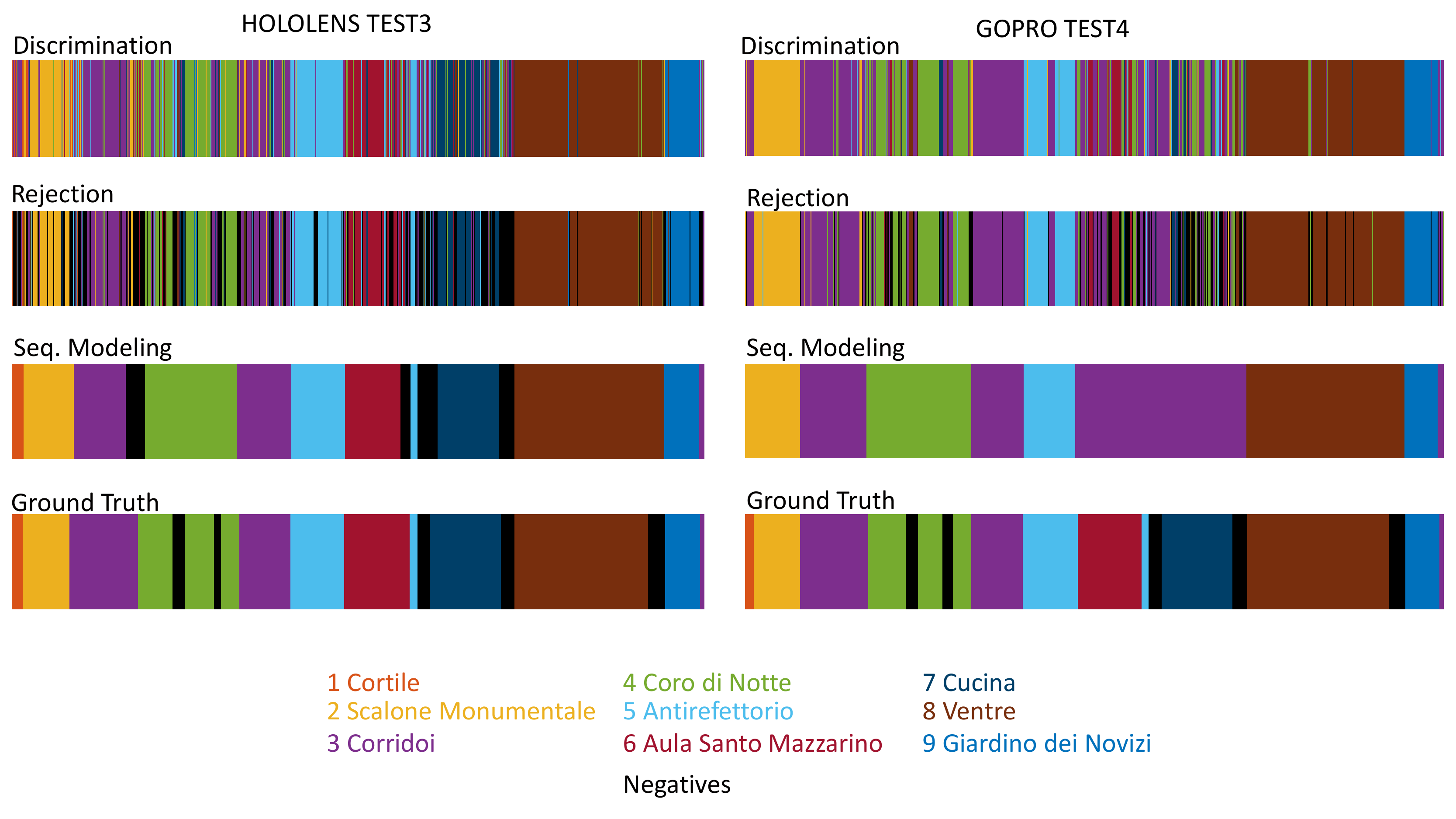}
	\caption{Color-coded segmentations for two corresponding test video acquired using HoloLens (left) and GoPro (right).}
	\label{fig:comparison}
\end{figure}

\figurename~\ref{fig:comparison} reports a qualitative comparison of the proposed method on "Test3" video (acquired by HoloLens) and "Test4" video (acquired by GoPro) used as validation videos. Please note that the two videos have been acquired simultaneously and so present similar content. 
The figure illustrates how the discrimination step allows to obtain more stable results in the case of GoPro data. For this reason, negative rejection tends to be more pronounced in the case of HoloLens data. The final segmentations obtained after the sequential modeling step are in general equivalent.

We would like to note that, even if the final results obtained using HoloLens and GoPro are equivalent in quantitative terms, the data acquired using the HoloLens device is deemed to carry more relevant information about what the user is actually looking at (see \figurename~\ref{fig:interest}). Such additional information can be leveraged in applications which go beyond localization, such as attention and behavioral modeling. Moreover, head-mounted devices such as HoloLens are better suited then chest-mounted cameras to provide additional services (e.g., augmented reality) to the visitor. This makes in our opinion HoloLens (and head-mounted devices in general) preferable. A series of demo videos to assess the performance of the investigated system are available at our web page \url{http://iplab.dmi.unict.it/VEDI/video.html}.

\begin{figure*}
\includegraphics[width=\linewidth]{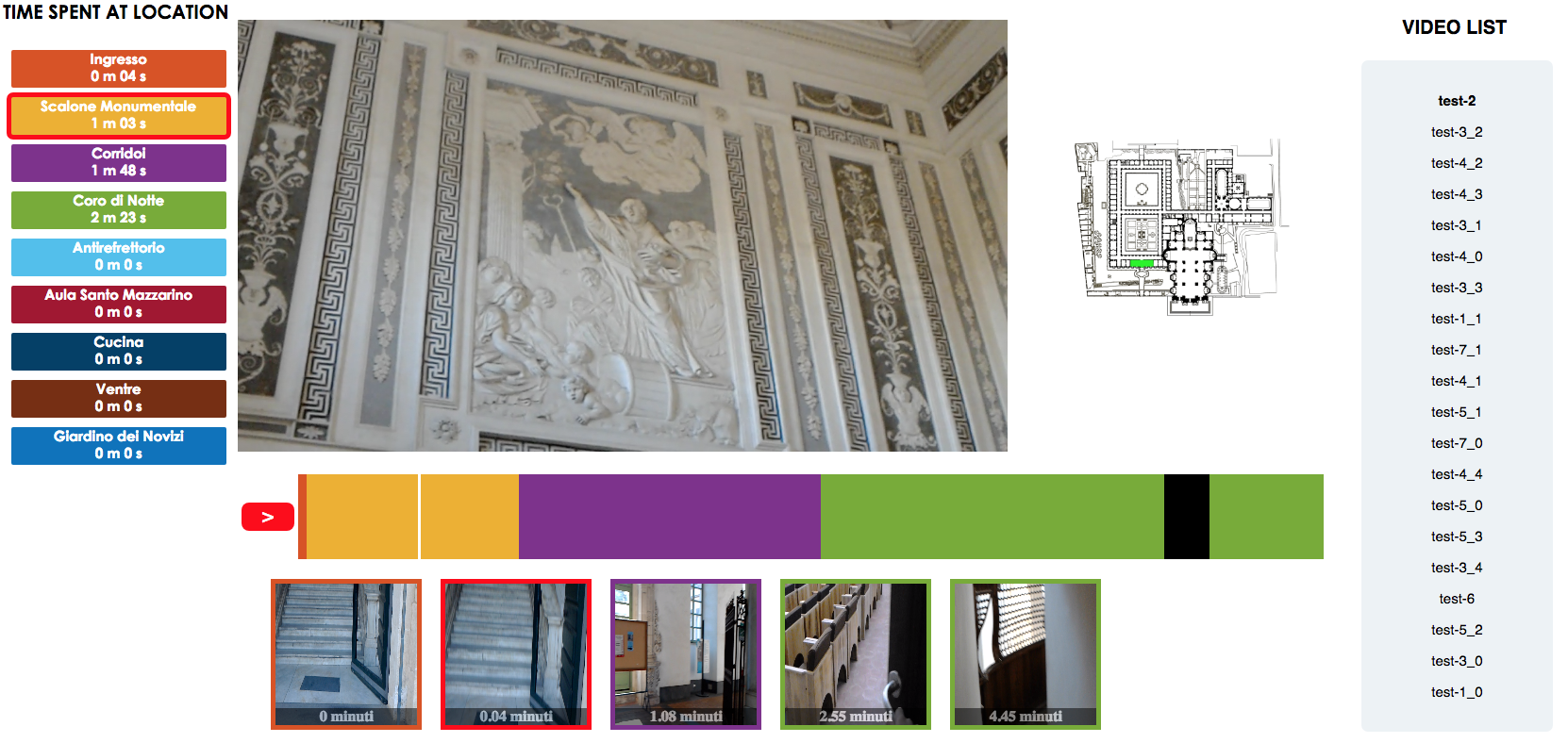}
\caption{The GUI to allow the site manager to analyze the captured videos.}
\label{fig:gui}
\end{figure*}

We also implemented a Manager Visualization Tool (MVT) as a web interface. The tool allows the manager to analyze the output of the system which automatically localizes the visitor in each frame of the video. The GUI allows the manager to select the video to be analyzed. A video player allows to skim through the video by clicking on a graphical representation of the inferred video segmentation. The GUI also presents the estimated time spent at each location and highlights the location of the visitor in a 2D map. Giving the opportunity to explore the video in a structured way, the system can provide useful information to the site manager. For instance, the site manager can infer which locations attract more the attention of the visitor, which can help improve the design of visit paths and profiling visitors. \figurename~\ref{fig:gui} illustrates the different components of the developed tool. A video demo of the developed interface is available at \url{http://iplab.dmi.unict.it/VEDI/demogui.html}.

\section{Conclusion}
\label{sec:conclusion}
This work has investigated the problem of localizing visitors in a cultural site using egocentric (first person) cameras. To study the problem of room-level localization, we proposed and publicly released a dataset containing more than 4 hours of egocentric video, labeled according to the location of the visitor and to the observed cultural object of interest. The localization problem has been investigated reporting baseline results using a state-of-the-art method on data acquired using a head-mounted HoloLens and a chest-mounted GoPro device. Despite the larger field of view of the GoPro device, HoloLens allows to achieve similar performance in the localization task. We believe that the proposed UNICT-VEDI dataset will encourage further research on this domain. Future works will consider the problem of understanding which cultural objects are observed by the visitors. This will allow to provide more detailed information on the behavior and preferences of the visitors to the site manager. Moreover, the analysis will be extended and generalized to other cultural sites. 
\section*{Acknowledgment}
This research is supported by PON MISE - Horizon 2020, Project VEDI - Vision Exploitation for Data Interpretation, Prog. n. F/050457/02/X32 - CUP: B68I17000800008 - COR: 128032, and Piano della Ricerca 2016-2018 linea di Intervento 2 of DMI of the University of Catania. We gratefully acknowledge the support of NVIDIA Corporation with the donation of the Titan X Pascal GPU used for this research.

\bibliographystyle{ACM-Reference-Format}
\bibliography{refs}

\pagebreak
\onecolumn
% !TeX spellcheck = en_GB
\section*{Supplementary Material}
\label{sec:supp}

The dataset has been acquired using two devices (HoloLens and GoPro) and hence comprises two device-specifc training/testing sets. Video frames are labelled according to: 1) the environment of the visitor, 2) the point of interest he is looking at. \figurename~\ref{fig:environments} reports some sample frames related to the different environments, whereas \figurename~\ref{fig:interest} reports some sample frames related to the different points of interest. \tablename~\ref{table:training_videos} summarizes the number of training videos acquired using HoloLens and GoPro devices. \tablename~\ref{table:testing_videos} summarizes the acquired test videos.
In the following sections, we report additional material related to the data acquired using the HoloLens and GoPro devices.

\begin{table}[t]
	\centering
	\vspace{1mm}
	\caption{Summary of the collected training videos. On the left, we report the number of videos collected for the points of interest. On the right, we report the informations for the videos of collected for the environments. The values in the \textit{Time (s)} and \textit{MB} columns are related to the average time and occupied memory respectively.}
	\vspace{-3mm}
	\label{table:training_videos}
	\begin{tabular}{|c|c|c|c|c|c|c|}
		\hline
		& \multicolumn{3}{c|}{\textbf{Points of interest}} & \multicolumn{3}{c|}{\textbf{Environments}} \\ \hline
		& \# Videos & Time (s) & MB     & \# Videos  & Time (s) & MB \\ \hline
		\textbf{Hololens} & 68                & 104           & 103             & 12            & 190       & 196        \\
		\textbf{GoPro}    & 68                & 93           & 233            & 12            & 186       & 455        \\ \hline
	\end{tabular}

\end{table}

\begin{table}[t]
	\centering

	\caption{Summary of the acquired test videos . The values on the \textit{Time} and \textit{MB} columns are average values.}
\label{table:testing_videos}	
\vspace{-3mm}
	\begin{tabular}{|c|c|c|c|}
		\hline
		& \multicolumn{3}{c|}{\textbf{Test Videos}} \\ \hline
		& Number of videos   & Time (s)   & Storage (MB)   \\ \hline
		\textbf{Hololens} & 7              & 780         & 730           \\
		\textbf{GoPro}    & 7              & 828         & 2062          \\ \hline
	\end{tabular}

\end{table}
\begin{table}[t]
	\centering
	\caption{List of training videos of environments acquired using Hololens. For each video we show: time, amount of occupied memory, number of frames, percentage of frames with respect to the total number of frames of the training set.}
	\label{table:Holo_Training}	
	
	\begin{tabular}{|l|c|c|c|c|}
		\hline
		\multicolumn{1}{|c|}{Name} & Time (s) & Storage (MB) & \#frames & \%frames \\ \hline
		1.0\_Cortile              & 48         & 48.145      & 1171    & 0,24\%  \\
		2.0\_Scalone              & 81         & 80.085      & 1947    & 0,40\%  \\
		2.0\_Scalone1             & 116        & 113.173     & 2747    & 0,56\%  \\
		2.0\_Scalone2             & 38         & 37.758      & 917     & 0,19\%  \\
		4.0\_CoroDiNotte          & 169        & 167.614     & 4068    & 0,83\%  \\
		4.0\_CoroDiNotte1         & 44         & 44.068      & 1067    & 0,22\%  \\
		5.0\_Antirefettorio       & 247        & 244.500     & 5933    & 1,21\%  \\
		5.0\_Antirefettorio1      & 263        & 260.395     & 6315    & 1,29\%  \\
		6.0\_SantoMazzarino       & 241        & 239.007     & 5800    & 1,18\%  \\
		7.0\_Cucina               & 239        & 237.268     & 5753    & 1,17\%  \\
		8.0\_Ventre               & 679        & 832.844     & 20385   & 4,15\%  \\
		9.0\_GiardinoNovizi       & 116        & 113.949     & 2788    & 0,57\%  \\ \hline
		AVG                     & 175.46     & 201.57      & 4908    & 1,00\%  \\ \hline
	\end{tabular}
	
\end{table} 

\subsection{HoloLens}
\tablename~\ref{table:Holo_Training} details the training videos acquired to represent each of the considered environments. For each class, we report the total duration of the videos, the required storage, the number of frames and the percentage of frames, with respect to the total number of frames in the training set.
\tablename~\ref{table:4} details the training videos acquired to represent each of the considered points of interest. For each class, we report the total duration of the videos, the required storage, the number of frames and the percentage of frames, with respect to the total number of frames in the training set.
\tablename~\ref{tab:test_hololens_details} reports a list of all test videos acquired using HoloLens. For each video, we report: Time, Storage, number of frames and percentage of frames with respect to the total number of frames of the training dataset.

\begin{longtable}[c]{|l|c|c|c|c|}	
\caption{\normalsize List of training videos of points of interest acquired with Hololens. For each video we show: Time, Storage, number of frames and percentage of frames with respect to the total number of frames of the training dataset. The table continues in the next page.}
\label{table:4}
\vspace{-3mm}
\\
\hline
\multicolumn{1}{|c|}{Name}             & Time (s) & Storage (MB) & \#frame & \%frame \\ \hline
\endfirsthead
\endhead
\hline
\endfoot
\endlastfoot
2.1\_Scalone\_RampaS.Nicola                & 163        & 161.776     & 3926    &      0,80\%
\\
2.2\_Scalone\_RampaS.Benedetto             & 14         & 14.471      & 354    &   0,07\%
\\
2.2\_Scalone\_RampaS.Benedetto1            & 148        & 147.226     & 3573    &    0,73\%
\\
3.1\_Corridoi\_TreBiglie              & 75         & 74.261      & 1804    &    0,37\%
\\
3.2\_Corridoi\_ChiostroLevante        & 55         & 54.500      & 1328    &   0,27\%
\\
3.3\_Corridoi\_Plastico               & 80         & 79.179      & 1927    &   0,39\%
\\
3.4\_Corridoi\_Affresco               & 74         & 74.023      & 1800    &   0,37\%
\\
3.5\_Corridoi\_Finestra\_ChiostroLev. & 84         & 83.102      & 2035    &    0,41\%
\\
3.6\_Corridoi\_PortaCoroDiNotte       & 53         & 53.255      & 1291    &  0,26\%
\\
3.6\_Corridoi\_PortaCoroDiNotte1      & 60         & 59.411      & 1446    &    0,29\%
\\
3.6\_Corridoi\_PortaCoroDiNotte2      & 58         & 57.901      & 1406    &   0,29\%
\\
3.7\_Corridoi\_TracciaPortone         & 53         & 53.118      & 1291    &     0,26\%
\\
3.8\_Corridoi\_StanzaAbate            & 81         & 79.773      & 1948    &     0,40\%
\\
3.9\_Corridoi\_CorridoioDiLevante     & 89         & 88.173      & 2142    &    0,44\%
\\
3.10\_Corridoi\_CorridoioCorodiNotte  & 122        & 121.321     & 2945    &     0,60\%
\\
3.10\_Corridoi\_CorridoioCorodiNotte1 & 103        & 101.796     & 2472    &   0,50\%
\\
3.11\_Corridoi\_CorridoioOrologio     & 300        & 296.488     & 7202    &    1,47\%
\\
4.1\_CoroDiNotte\_Quadro              & 71         & 70.245      & 1716    &   0,35\%
\\
4.2\_CoroDiNotte\_Pav.Orig.Altare           & 73         & 72.358      & 1759    &    0,36\%
\\
4.3\_CoroDiNotte\_BalconeChiesa            & 39         & 39.331      & 957     &    0,19\%
\\
4.3\_CoroDiNotte\_BalconeChiesa1            & 49         & 48.550      & 1185    &     0,24\%
\\
4.3\_CoroDiNotte\_BalconeChiesa2            & 80         & 79.861      & 1935    &    0,39\%
\\
5.1\_Antirefettorio\_PortaA.S.Mazz.Ap.    & 67         & 67.001      & 1630    &    0,33\%
\\
5.1\_Antirefettorio\_PortaA.S.Mazz.Ch.     & 74         & 73.589      & 1785    &    0,36\%
\\
5.2\_Antirefettorio\_PortaMuseoFab.Ap.     & 50         & 49.840      & 1211    &    0,25\%
\\
5.2\_Antirefettorio\_PortaMuseoFab.Ch.     & 62         & 61.846      & 1503    &   0,31\%
\\
5.3\_Antirefettorio\_PortaAntiref.           & 51         & 58.557      & 1537    &    0,31\%
\\
5.4\_Antirefettorio\_PortaRef.Piccolo           & 54         & 53.767      & 1306    &   0,27\%
\\
5.5\_Antirefettorio\_Cupola           & 57         & 56.089      & 1377    &   0,28\%
\\
5.6\_Antirefettorio\_AperturaPavimento        & 55         & 54.586      & 1322    &   0,27\%
\\
5.7\_Antirefettorio\_S.Agata          & 48         & 47.820      & 1165    &    0,24\%
\\
5.8\_Antirefettorio\_S.Scolastica           & 58         & 57.919      & 1407    &    0,29\%
\\
5.9\_Antirefettorio\_ArcoconFirma             & 62         & 76.700      & 1864    &    0,38\%
\\
5.10\_Antirefettorio\_BustoVaccarini           & 65         & 64.298      & 1563    &    0,32\%
\\
6.1\_SantoMazzarino\_QuadroS.Mazz.          & 71         & 70.401      & 1716    &     0,35\%
\\
6.2\_SantoMazzarino\_Affresco         & 213        & 211.424     & 5124    &   1,04\%
\\
6.3\_SantoMazzarino\_PavimentoOr.     & 99         & 98.691      & 2397    &     0,49\%
\\
6.4\_SantoMazzarino\_PavimentoRes.    & 69         & 69.148      & 1675    &    0,34\%
\\
6.5\_SantoMazzarino\_BassorilieviManc.     & 117        & 115.348     & 2823    &    0,57\%
\\
6.6\_SantoMazzarino\_LavamaniSx       & 151        & 149.882     & 3637    &     0,74\%
\\
6.7\_SantoMazzarino\_LavamaniDx       & 93         & 92.928      & 2256    &    0,46\%
\\
6.8\_SantoMazzarino\_TavoloRelatori      & 150        & 148.661     & 3603    &    0,73\%
\\
6.9\_SantoMazzarino\_Poltrone         & 108        & 107.374     & 2604    &    0,53\%
\\
7.1\_Cucina\_Edicola                  & 369        & 437.219     & 11086   &    2,26\%
\\
7.2\_Cucina\_PavimentoA               & 52         & 52.163      & 1268    &     0,26\%
\\
7.3\_Cucina\_PavimentoB               & 52         & 52.244      & 1266    &     0,26\%
\\
7.4\_Cucina\_PassavivandePavim.Orig.             & 81         & 80.733      & 1961    &     0,40\%
\\
7.5\_Cucina\_AperturaPav.1                    & 57         & 57.170      & 1385    &     0,28\%
\\
7.5\_Cucina\_AperturaPav.2                    & 53         & 52.875      & 1280    &     0,26\%
\\
7.5\_Cucina\_AperturaPav.3                    & 62         & 62.320      & 1509    &     0,31\%
\\
7.6\_Cucina\_Scala                    & 77         & 76.587      & 1856    &     0,38\%
\\
7.7\_Cucina\_SalaMetereologica         & 156        & 154.394     & 3748    &   0,76\%
\\
8.1\_Ventre\_Doccione                 & 103        & 102.683     & 2492    &      0,51\%
\\
8.2\_Ventre\_VanoRacc.Cenere               & 126        & 124.837     & 3026    &     0,62\%
\\
8.3\_Ventre\_SalaRossa                & 300        & 296.792     & 7202    &     1,47\%
\\
8.4\_Ventre\_ScalaCucina              & 214        & 212.372     & 5152    &     1,05\%
\\
8.5\_Ventre\_CucinaProvv.             & 148        & 146.379     & 3553    &    0,72\%
\\
8.6\_Ventre\_Ghiacciaia               & 69         & 68.562      & 1668    &      0,34\%
\\
8.6\_Ventre\_Ghiacciaia1              & 266        & 263.817     & 6398    &    1,30\%
\\
8.7\_Ventre\_Latrina                    & 102        & 100.542     & 2468    &     0,50\%
\\
8.8\_Ventre\_OssaScarti                     & 154        & 152.861     & 3713    &     0,76\%
\\
8.8\_Ventre\_OssaScarti1                    & 36         & 36.345      & 886     &    0,18\%
\\
8.9\_Ventre\_Pozzo                    & 300        & 297.167     & 7209    &    1,47\%
\\
8.10\_Ventre\_Cisterna                & 57         & 56.572      & 1384    &      0,28\%
\\
8.11\_Ventre\_BustoP.Tacchini           & 110        & 109.520     & 2662    &    0,54\%
\\
9.1\_GiardinoNovizi\_NicchiaePav.     & 106        & 105.004     & 2555    &     0,52\%
\\
9.2\_GiardinoNovizi\_TraccePalestra  & 63         & 62.464      & 1525    &   0,31\%
\\
9.3\_GiardinoNovizi\_Pergolato           & 166        & 164.150     & 3984    &     0,81\%
\\ \hline
AVG                                 & 102.6      & 103.26      & 2517    &   0,51\%
\\ \hline

\end{longtable}

\begin{table}[]
	\caption{List of test videos acquired using HoloLens. For each video we report the length in seconds, the amount of occupied memory, the number of frames, the number of environments included in the video, the number of points of interest included in the video, the sequence of environments as navigated by the visitor, and the sequence of points of interest, as navigated by the visitor.}
	\label{tab:test_hololens_details}
	\centering
	\resizebox{\textwidth}{!}{%
		\begin{tabular}{|l|c|c|c|c|c|c|l|l|}
			\hline
			Video    & Time                     & MB      & \multicolumn{1}{l|}{\#frames} & \multicolumn{1}{l|}{\%frames} & \multicolumn{1}{l|}{\#env.} & \multicolumn{1}{l|}{\#p.int.} & seq.environments                                                            & seq. p.interest                                                                                                                                                                                                                                                                                                                                                      \\ \hline
			Test1   & 300                      & 296.896 & 7202                         & 5,49\%                       & 4                               & 10                                 & 1-\textgreater2-\textgreater3-\textgreater4                             & \begin{tabular}[c]{@{}l@{}}1.1-\textgreater2.2-\textgreater3.1-\textgreater3.9-\textgreater3.10-\textgreater3.4-\textgreater\\ 3.6-\textgreater4.1-\textgreater4.2-\textgreater4.3\end{tabular}                                                                                                                                                                       \\
			Test1.1 & 300                      & 297.031 & 7202                         & 5,49\%                       & 4                               & 11                                 & 4-\textgreater3-\textgreater5-\textgreater6                             & \begin{tabular}[c]{@{}l@{}}4.3-\textgreater3.11-\textgreater5.6-\textgreater5.7-\textgreater5.1-\textgreater6.1-\textgreater\\ 6.4-\textgreater6.9-\textgreater6.3-\textgreater6.6-\textgreater6.2\end{tabular}                                                                                                                                                       \\ \hline
			Test2.0 & 300                      & 296.993 & 7203                         & 5,49\%                       & 4                               & 9                                  & 1\_\textgreater2-\textgreater3-\textgreater4                            & \begin{tabular}[c]{@{}l@{}}1.1-\textgreater2.1-\textgreater3.9-\textgreater3.10-\textgreater3.5-\textgreater3.10-\textgreater\\ 3.5-\textgreater3.10-\textgreater3.6-\textgreater4.1-\textgreater4.2-\textgreater4.3\end{tabular}                                                                                                                                     \\ \hline
			Test3.0 & 300                      & 296.959 & 7202                         & 5,49\%                       & 4                               & 8                                  & 1-\textgreater2-\textgreater3-\textgreater4                             & \begin{tabular}[c]{@{}l@{}}1.1-\textgreater2.1-\textgreater3.9-\textgreater3.10-\textgreater3.4-\textgreater3.11-\textgreater\\ 3.6-\textgreater4.1\end{tabular}                                                                                                                                                                                                      \\
			Test3.1 & 300                      & 296.913 & 7203                         & 5,49\%                       & 3                               & 5                                  & 4-\textgreater3-\textgreater5                                           & 4.2-\textgreater4.3-\textgreater4.2-\textgreater3.11-\textgreater5.5-\textgreater5.6                                                                                                                                                                                                                                                                                  \\
			Test3.2 & 300                      & 297.083 & 7201                         & 5,49\%                       & 4                               & 17                                 & 5-\textgreater6-\textgreater5-\textgreater7                             & \begin{tabular}[c]{@{}l@{}}5.6-\textgreater5.1-\textgreater5.3-\textgreater5.10-\textgreater5.9-\textgreater5.1-\textgreater\\ 6.2-\textgreater6.1-\textgreater6.4-\textgreater6.9-\textgreater6.5-\textgreater6.4-\textgreater\\ 6.3-\textgreater6.8-\textgreater6.5-\textgreater6.7-\textgreater6.6-\textgreater6.5-\textgreater\\ 5.2-\textgreater7.1\end{tabular} \\
			Test3.3 & 300                      & 297.160 & 7202                         & 5,49\%                       & 2                               & 12                                 & 7-\textgreater8                                                         & \begin{tabular}[c]{@{}l@{}}7.2.\textgreater7.6-\textgreater7.5-\textgreater7.4-\textgreater7.7-\textgreater7.1-\textgreater\\ 8.1-\textgreater8.6-\textgreater8.2-\textgreater8.8-\textgreater8.4-\textgreater8.10\end{tabular}                                                                                                                                       \\
			Test3.4 & 237                      & 234.744 & 5694                         & 4,34\%                       & 3                               & 5                                  & 8-\textgreater9-\textgreater3                                           & 8.9-\textgreater8.11-\textgreater8.3-\textgreater9.1-\textgreater9.2                                                                                                                                                                                                                                                                                                  \\ \hline
			Test4.0 & 300                      & 296.979 & 7204                         & 5,49\%                       & 4                               & 10                                 & 
			1-\textgreater2-\textgreater3-\textgreater5                             &
			\begin{tabular}[c]{@{}l@{}}
			1.1-\textgreater2.2-\textgreater3.9-\textgreater3.10-\textgreater3.4-\textgreater3.11-\textgreater\\-\textgreater3.7-\textgreater3.11-\textgreater5.10-\textgreater5.9-\textgreater5.1 
			\end{tabular} 
			                                                                                               \\
			Test4.1 & 300                      & 296.654 & 7202                         & 5,49\%                       & 3                               & 16                                 & 5-\textgreater7-\textgreater8                                           & \begin{tabular}[c]{@{}l@{}}5.6-\textgreater5.5-\textgreater5.8-\textgreater5.4-\textgreater5.2-\textgreater7.1-\textgreater\\ 7.3-\textgreater7.2-\textgreater7.1-\textgreater7.4-\textgreater7.1-\textgreater7.5-\textgreater\\ 7.7-\textgreater7.1-\textgreater8.1-\textgreater8.5-\textgreater8.2-\textgreater8.4-\textgreater\\ 8.7\end{tabular}                  \\
			Test4.2 & 136                      & 135.143 & 3281                         & 2,50\%                       & 1                               & 3                                  & 8                                                                       & 8.10-\textgreater8.9-\textgreater8.11                                                                                                                                                                                                                                                                                                                                 \\
			Test4.3 & 300                      & 296.853 & 7202                         & 5,49\%                       & 4                               & 7                                  & 8-\textgreater9-\textgreater3-\textgreater4                             & \begin{tabular}[c]{@{}l@{}}8.3-\textgreater9.1-\textgreater9.3-\textgreater9.2-\textgreater3.11-\textgreater3.6-\textgreater\\ 4.2\end{tabular}                                                                                                                                                                                                                       \\
			Test4.4 & 201                      & 199.585 & 4845                         & 3,69\%                       & 3                               & 4                                  & 4-\textgreater3-\textgreater2                                           & 4.2-\textgreater3.10-\textgreater3.9-\textgreater2.1                                                                                                                                                                                                                                                                                                                  \\ \hline
			Test5.0 & 274                      & 271.524 & 6590                         & 5,02\%                       & 4                               & 11                                 & 1-\textgreater2-\textgreater3-\textgreater2-\textgreater3-\textgreater4 & \begin{tabular}[c]{@{}l@{}}1.1-\textgreater3.1-\textgreater3.3-\textgreater3.2-\textgreater2.1-\textgreater3.9-\textgreater\\ 3.10-\textgreater3.5-\textgreater3.10-\textgreater3.5-\textgreater3.10-\textgreater3.4-\textgreater\\ 3.7-\textgreater3.4-\textgreater3.6\end{tabular}                                                                                  \\
			Test5.1 & 300                      & 296.801 & 7202                         & 5,49\%                       & 4                               & 11                                 & 4-\textgreater3-\textgreater9-\textgreater3-\textgreater5               & 
			\begin{tabular}[c]{@{}l@{}}
				4.1-\textgreater4.2-\textgreater3.6-\textgreater3.11-\textgreater9.2-\textgreater3.11-\textgreater\\-\textgreater5.5-\textgreater5.3-\textgreater5.9-\textgreater5.7-\textgreater5.8-\textgreater5.2                                              
			\end{tabular} 
		
			                                                                                                                                   \\
			Test5.2 & 300                      & 296.921 & 7202                         & 5,49\%                       & 2                               & 13                                 & 7-\textgreater8                                                         & \begin{tabular}[c]{@{}l@{}}7.1-\textgreater7.2-\textgreater7.3-\textgreater7.1-\textgreater7.5-\textgreater7.1-\textgreater\\ 7.1-\textgreater7.4-\textgreater7.6-\textgreater7.7-\textgreater7.4-\textgreater7.7-\textgreater\\ 8.1-\textgreater8.5-\textgreater8.4-\textgreater8.5-\textgreater8.6-\textgreater8.2-\textgreater\\ 8.8-\textgreater8.4\end{tabular}  \\
			Test5.3 & 300                      & 297.063 & 7201                         & 5,49\%                       & 4                               & 17                                 & 8-\textgreater7-\textgreater5-\textgreater6                             & \begin{tabular}[c]{@{}l@{}}8.11-\textgreater8.9-\textgreater8.3-\textgreater8.4-\textgreater8.5-\textgreater8.6-\textgreater\\ 7.1-\textgreater7.5-\textgreater7.2-\textgreater5.1-\textgreater6.2-\textgreater6.4-\textgreater\\ 6.1-\textgreater6.5-\textgreater6.3-\textgreater6.5-\textgreater6.2-\textgreater6.7-\textgreater\\ 6.9\end{tabular}                 \\ \hline
			Test6.0 & \multicolumn{1}{l|}{300} & 296.880 & 7202                         & 5,49\%                       & 3                               & 9                                  & 1-\textgreater2-\textgreater3                                           & \begin{tabular}[c]{@{}l@{}}1.1-\textgreater2.2-\textgreater2.1-\textgreater3.9-\textgreater3.10-\textgreater3.4-\textgreater\\ 3.11-\textgreater3.6-\textgreater3.11-\textgreater3.7-\textgreater3.11\end{tabular}                                                                                                                                                    \\ \hline
			Test7.0 & \multicolumn{1}{l|}{300} & 296.945 & 7202                         & 5,49\%                       & 3                               & 7                                  & 1-\textgreater2-\textgreater3                                           & \begin{tabular}[c]{@{}l@{}}1.1-\textgreater2.2-\textgreater3.9-\textgreater3.10-\textgreater3.4-\textgreater3.6-\textgreater\\ 3.11\end{tabular}                                                                                                                                                                                                                      \\
			Test7.1 & \multicolumn{1}{l|}{113} & 112.030 & 2721                         & 2,07\%                       & 3                               & 3                                  & 3-\textgreater5-\textgreater3                                           & 3.11-\textgreater5.9-\textgreater5.10-\textgreater3.11                   \\
			\hline                                                                                                                                                                                                                                                                                             
		\end{tabular}%
	}
\end{table}

\clearpage
\subsection{GoPro}
\tablename~\ref{table:GoPro_Training} details the training videos acquired to represent each of the considered environments. For each class, we report the total duration of the videos, the required storage, the number of frames and the percentage of frames, with respect to the total number of frames in the training set.
\tablename~\ref{table:5} details the training videos acquired to represent each of the considered points of interest. For each class, we report the total duration of the videos, the required storage, the number of frames and the percentage of frames, with respect to the total number of frames in the training set.
\tablename~\ref{tab:test_gopro_details} reports a list of all test videos acquired using GoPro. For each video, we report: Time, Storage, number of frames and percentage of frames with respect to the total number of frames of the training dataset.

\begin{table}[t]
	\centering
	\caption{List of training videos of environments acquired using GoPro. For each video we show: time, amount of occupied memory, number of frames, percentage of frames with respect to the total number of frames of the training set.}
	\label{table:GoPro_Training}
	\begin{tabular}{|l|c|c|c|c|}
		\hline
		\multicolumn{1}{|c|}{Name} & Time (s) & Storage (MB) & \#frames & \%frames \\ \hline
		1.0\_Cortile              & 47         & 116.798     & 1187    & 0,24\%  \\
		2.0\_Scalone              & 81         & 201.194     & 2044    & 0,42\%  \\
		2.0\_Scalone1             & 111        & 274.094     & 2785    & 0,57\%  \\
		2.0\_Scalone2             & 36         & 90.250      & 918     & 0,19\%  \\
		4.0\_CoroDiNotte          & 168        & 413.702     & 4205    & 0,86\%  \\
		4.0\_CoroDiNotte1         & 43         & 107.968     & 1097    & 0,22\%  \\
		5.0\_Antirefettorio       & 191        & 471.275     & 4790    & 0,97\%  \\
		5.0\_Antirefettorio1      & 262        & 644.852     & 6554    & 1,33\%  \\
		6.0\_SantoMazzarino       & 240        & 591.177     & 6009    & 1,22\%  \\
		7.0\_Cucina               & 241        & 592.907     & 6027    & 1,23\%  \\
		8.0\_Ventre               & 699        & 1719.352    & 17478   & 3,56\%  \\
		9.0\_GiardinoNovizi       & 116        & 427.472     & 6965    & 1,42\%  \\ \hline
		AVG                     & 186.25     & 242.62      & 5005    & 1,02\%  \\ \hline
	\end{tabular}
\end{table}

\begin{longtable}[c]{|l|c|c|c|c|}
	\caption{\normalsize List of training videos of environments acquired using GoPro. For each video we show: time, amount of occupied memory, number of frames, percentage of frames with respect to the total number of frames of the training set.}
	\label{table:5}
	\vspace{-3mm}
	\\
	\hline
	\multicolumn{1}{|c|}{Name}             & Time (s) & Storage (MB) & \#frame & \%frame \\ \hline
	\endfirsthead
	\endhead
	\hline
	\endfoot
	\endlastfoot
	2.1\_Scalone\_RampaS.Nicola                & 160        & 394,844     & 4013    &     0,82\%
	\\
	2.2\_Scalone\_RampaS.Benedetto             & 14         & 35,758      & 363     &      0,07\%
	\\
	2.2\_Scalone\_RampaS.Benedetto1            & 147        & 361,856     & 3678    &     0,75\%
	\\
	3.1\_Corridoi\_TreBiglie              & 72         & 178,313     & 1812    &   0,37\%
	\\
	3.2\_Corridoi\_ChiostroLevante        & 54         & 202,165     & 3295    &    0,67\%
	\\
	3.3\_Corridoi\_Plastico               & 79         & 195,673     & 1989    &    0,40\%
	\\
	3.4\_Corridoi\_Affresco               & 74         & 182,106     & 1850    &    0,38\%
	\\
	3.5\_Corridoi\_Finestra\_ChiostroLev. & 81         & 300,474     & 4896    &     1,00\%
	\\
	3.6\_Corridoi\_PortaCoroDiNotte       & 50         & 123,34      & 1253    &     0,26\%
	\\
	3.6\_Corridoi\_PortaCoroDiNotte1      & 60         & 148,97      & 1514    &    0,31\%
	\\
	3.6\_Corridoi\_PortaCoroDiNotte2      & 58         & 142,747     & 1451    &    0,30\%
	\\
	3.7\_Corridoi\_TracciaPortone         & 55         & 136,484     & 1387    &    0,28\%
	\\
	3.8\_Corridoi\_StanzaAbate            & 58         & 122,859     & 1755    &     0,36\%
	\\
	3.9\_Corridoi\_CorridoioDiLevante     & 85         & 211,539     & 2148    &   0,44\%
	\\
	3.10\_Corridoi\_CorridoioCorodiNotte  & 120        & 297,325     & 3022    &     0,62\%
	\\
	3.10\_Corridoi\_CorridoioCorodiNotte1 & 100        & 246,769     & 2509    &   0,51\%
	\\
	3.11\_Corridoi\_CorridoioOrologio     & 176        & 434,081     & 4412    &   0,90\%
	\\
	4.1\_CoroDiNotte\_Quadro              & 69         & 171,551     & 1743    &    0,35\%
	\\
	4.2\_CoroDiNotte\_Pav.Orig.Altare           & 72         & 178,448     & 1813    &   0,37\%
	\\
	4.3\_CoroDiNotte\_BalconeChiesa             & 39         & 96,281      & 978     &    0,20\%
	\\
	4.3\_CoroDiNotte\_BalconeChiesa1            & 48         & 119,512     & 1214    &    0,25\%
	\\
	4.3\_CoroDiNotte\_BalconeChiesa2            & 80         & 197,29      & 2004    &    0,41\%
	\\
	5.1\_Antirefettorio\_PortaA.S.Mazz.Ap.     & 66         & 164,644     & 1673    &    0,34\%
	\\
	5.1\_Antirefettorio\_PortaA.S.Mazz.Ch     & 74         & 183,481     & 1864    &    0,38\%
	\\
	5.2\_Antirefettorio\_PortaMuseoFab.Ap     & 50         & 124,991     & 1270    &    0,26\%
	\\
	5.2\_Antirefettorio\_PortaMuseoFab.Ch     & 62         & 154,254     & 1567    &     0,32\%
	\\
	5.3\_Antirefettorio\_PortaAntiref.    & 52         & 128,695     & 1306    &     0,27\%
	\\
	5.4\_Antirefettorio\_PortaRef.Piccolo       & 54         & 133,494     & 1356    &     0,28\%
	\\
	5.5\_Antirefettorio\_Cupola           & 55         & 135,594     & 1378    &     0,28\%
	\\
	5.6\_Antirefettorio\_AperturaPavimento        & 57         & 141,647     & 1439    &     0,29\%
	\\
	5.7\_Antirefettorio\_S.Agata          & 49         & 120,611     & 1225    &     0,25\%
	\\
	5.8\_Antirefettorio\_S.Scolastica           & 56         & 137,79      & 1400    &     0,28\%
	\\
	5.9\_Antirefettorio\_ArcoconFirma             & 60         & 149,889     & 1522    &     0,31\%
	\\
	5.10\_Antirefettorio\_BustoVaccarini          & 66         & 162,799     & 1654    &    0,34\%
	\\
	6.1\_SantoMazzarino\_QuadroS.Mazz.           & 72         & 178,16      & 1810    &    0,37\%
	\\
	6.2\_SantoMazzarino\_Affresco         & 214        & 526,612     & 5353    &    1,09\%
	\\
	6.3\_SantoMazzarino\_PavimentoOr.     & 98         & 242,332     & 2462    &   0,50\%
	\\
	6.4\_SantoMazzarino\_PavimentoRes.    & 68         & 169,524     & 1722    &     0,35\%
	\\
	6.5\_SantoMazzarino\_BassorilieviMan.     & 116        & 25,942      & 2906    &    0,59\%
	\\
	6.6\_SantoMazzarino\_LavamaniSx       & 134        & 331,517     & 3369    &     0,69\%
	\\
	6.7\_SantoMazzarino\_LavamaniDx       & 91         & 225,892     & 2296    &      0,47\%
	\\
	6.8\_SantoMazzarino\_TavoloRelatori      & 149        & 366,575     & 3726    &     0,76\%
	\\
	6.9\_SantoMazzarino\_Poltrone         & 108        & 266,611     & 2709    &     0,55\%
	\\
	7.1\_Cucina\_Edicola                  & 368        & 931,368     & 9467    &    1,93\%
	\\
	7.2\_Cucina\_PavimentoA               & 55         & 137,197     & 1394    &      0,28\%
	\\
	7.3\_Cucina\_PavimentoB               & 52         & 129,601     & 1316    &    0,27\%
	\\
	7.4\_Cucina\_PassavivandePavim.Orig.             & 84         & 206,667     & 2100    &    0,43\%
	\\
	7.5\_Cucina\_AperturaPav.1                    & 58         & 142,7       & 1450    &      0,30\%
	\\
	7.5\_Cucina\_AperturaPav.2                    & 57         & 142,145     & 1444    &     0,29\%
	\\
	7.5\_Cucina\_AperturaPav.3                    & 65         & 161,61      & 1641    &     0,33\%
	\\
	7.6\_Cucina\_Scala                    & 79         & 195,547     & 1988    &    0,40\%
	\\
	7.7\_Cucina\_SalaMetereologica         & 158        & 390,583     & 3970    &   0,81\%
	\\
	8.1\_Ventre\_Doccione                 & 100        & 246,353     & 2503    &     0,51\%
	\\
	8.2\_Ventre\_VanoRacc.Cenere               & 128        & 315,824     & 3210    &     0,65\%
	\\
	8.3\_Ventre\_SalaRossa                & 158        & 388,603     & 3951    &    0,80\%
	\\
	8.4\_Ventre\_ScalaCucina              & 213        & 524,06      & 5327    &    1,08\%
	\\
	8.5\_Ventre\_CucinaProvv.             & 151        & 371,84      & 3779    &      0,77\%
	\\
	8.6\_Ventre\_Ghiacciaia               & 72         & 178,295     & 1811    &     0,37\%
	\\
	8.6\_Ventre\_Ghiacciaia1              & 185        & 457,1       & 4646    &      0,95\%
	\\
	8.7\_Ventre\_Latrina                    & 80         & 295,294     & 4804    &    0,98\%
	\\
	8.8\_Ventre\_OssaScarti                     & 150        & 369,523     & 3756    &     0,76\%
	\\
	8.8\_Ventre\_OssaScarti1                    & 34         & 86,105      & 873     &    0,18\%
	\\
	8.9\_Ventre\_Pozzo                    & 149        & 367,713     & 3737    &     0,76\%
	\\
	8.10\_Ventre\_Cisterna                & 30         & 114,114     & 1852    &      0,38\%
	\\
	8.11\_Ventre\_BustoP.Tacchini           & 106        & 261,822     & 2661    &     0,54\%
	\\
	9.1\_GiardinoNovizi\_NicchiaePav.     & 104        & 256,731     & 2609    &     0,53\%
	\\
	9.2\_GiardinoNovizi\_TraccePalestra  & 58         & 216,801     & 3533    &      0,72\%
	\\
	9.3\_GiardinoNovizi\_Pergolato          & 164        & 404,893     & 4115    &      0,84\%
	\\ \hline
	AVG                                 & 93.53      & 232.97      & 2515    &      0,51\%
	\\ \hline
	
\end{longtable}

\clearpage
\begin{table}[t]
	\caption{List of test videos acquired using GoPro. For each video we report the length in seconds, the amount of occupied memory, the number of frames, the number of environments included in the video, the number of points of interest included in the video, the sequence of environments as navigated by the visitor, and the sequence of points of interest, as navigated by the visitor.}
	\label{tab:test_gopro_details}
	\centering
	\resizebox{\textwidth}{!}{%
		\begin{tabular}{|l|c|c|c|c|c|c|l|l|}
			\hline
			Video  & Time(s) & MB       & \multicolumn{1}{l|}{\#frames} & \multicolumn{1}{l|}{\%frames} & \multicolumn{1}{l|}{\#env.} & \multicolumn{1}{l|}{\#p.int.} & sequence environements                                                                                                                                                                                                                                                                                                                                                                         & sequence points of interest                                                                                                                                                                                                                                                                                                                                                                                                                                                                                                                                                                                                                                                                                                                                                                                                                                                                                                                                                                                                                                                                                                                                                                                                                                                            \\ \hline
			Test1 & 397  & 390.558  & 14788                        & 5.32\%                       & 4                               & 9                                  & 1.0-\textgreater2.0-\textgreater3.0-\textgreater4.0-\textgreater3.0                                                                                                                                                                                                                                                                                                                  & \begin{tabular}[c]{@{}l@{}}1.1-\textgreater2.2-\textgreater3.9-\textgreater3.10-\textgreater3.6-\textgreater4.1-\textgreater\\ 4.2-\textgreater4.3-\textgreater3.11\end{tabular}                                                                                                                                                                                                                                                                                                                                                                                                                                                                                                                                                                                                                                                                                                                                                                                                                                                                                                                                                                                                                                                                                            \\ \hline
			Test2 & 420  & 1033.149 & 10503                        & 3.78\%                       & 4                               & 8                                  & 1.0-\textgreater2.0-\textgreater3.0-\textgreater5.0-\textgreater3.0                                                                                                                                                                                                                                                                                                                  & \begin{tabular}[c]{@{}l@{}}1.1-\textgreater2.1-\textgreater3.9-\textgreater3.10-\textgreater3.4-\textgreater3.11-\textgreater\\ 3.6-\textgreater3.11-\textgreater5.10-\textgreater3.10\end{tabular}                                                                                                                                                                                                                                                                                                                                                                                                                                                                                                                                                                                                                                                                                                                                                                                                                                                                                                                                                                                                                                                                         \\ \hline
			Test3 & 579  & 1425.480 & 14491                        & 5.21\%                       & 5                               & 11                                 & 1.0-\textgreater2.0-\textgreater3.0-\textgreater9.0-\textgreater3.0-\textgreater5.0-\textgreater3.0                                                                                                                                                                                                                                                                                  & \begin{tabular}[c]{@{}l@{}}1.1-\textgreater2.2-\textgreater2.1-\textgreater2.2-\textgreater3.9-\textgreater3.10-\textgreater\\ 3.6-\textgreater3.11-\textgreater9.2-\textgreater9.1-\textgreater9.3-\textgreater9.2-\textgreater\\ 3.11-\textgreater5.10-\textgreater3.11\end{tabular}                                                                                                                                                                                                                                                                                                                                                                                                                                                                                                                                                                                                                                                                                                                                                                                                                                                                                                                                                                                      \\ \hline
			Test4 & 1472 & 3620.627 & 36808                        & 13.24\%                      & 9                               & 35                                 & \begin{tabular}[c]{@{}l@{}}1.0-\textgreater2.0-\textgreater3.0-\textgreater4.0-\textgreater3.0-\textgreater5.0-\textgreater\\ 6.0-\textgreater5.0-\textgreater7.0-\textgreater8.0-\textgreater9.0-\textgreater3.0\end{tabular}                                                                                                                                                       & \begin{tabular}[c]{@{}l@{}}1.1-\textgreater2.1-\textgreater2.2-\textgreater3.9-\textgreater3.10-\textgreater3.4-\textgreater\\ 3.11-\textgreater3.6-\textgreater4.1-\textgreater4.3-\textgreater3.11-\textgreater5.4-\textgreater\\ 5.2-\textgreater5.1-\textgreater5.3-\textgreater5.10-\textgreater5.9-\textgreater5.6-\textgreater5.1-\textgreater\\ 6.9-\textgreater6.4-\textgreater6.9-\textgreater6.8-\textgreater5.3-\textgreater5.6-\textgreater\\ 5.4-\textgreater5.2-\textgreater7.1-\textgreater7.6-\textgreater7.1-\textgreater7.5-\textgreater\\ 7.4-\textgreater7.7-\textgreater8.1-\textgreater8.6-\textgreater8.2-\textgreater8.8-\textgreater\\ 8.1-\textgreater8.10-\textgreater8.9-\textgreater8.11-\textgreater8.3-\textgreater9.1-\textgreater9.2\end{tabular}                                                                                                                                                                                                                                                                                                                                                                                                                                                                                         \\ \hline
			Test5 & 751  & 1848.142 & 18788                        & 6.76\%                       & 6                               & 26                                 & 1.0-\textgreater2.0-\textgreater3.0-\textgreater5.0-\textgreater7.0-\textgreater8.0                                                                                                                                                                                                                                                                                                  & \begin{tabular}[c]{@{}l@{}}1.1-\textgreater2.2-\textgreater2.1-\textgreater3.1-\textgreater3.9-\textgreater3.10-\textgreater\\ 3.4-\textgreater3.11-\textgreater5.10-\textgreater5.6-\textgreater5.1-\textgreater5.3-\textgreater\\ 5.4-\textgreater5.2-\textgreater7.1-\textgreater7.4-\textgreater7.1-\textgreater7.5-\textgreater\\ 7.7-\textgreater8.1-\textgreater8.5-\textgreater8.4-\textgreater8.7-\textgreater8.10-\textgreater\\ 8.8-\textgreater8.10-\textgreater8.11\end{tabular}                                                                                                                                                                                                                                                                                                                                                                                                                                                                                                                                                                                                                                                                                                                                                                               \\ \hline
			Test6 & 506  & 1245.254 & 12661                        & 4.56\%                       & 5                               & 11                                 & 8.0-\textgreater9.0-\textgreater3.0-\textgreater4.0-\textgreater3.0-\textgreater2.0                                                                                                                                                                                                                                                                                                  & \begin{tabular}[c]{@{}l@{}}8.3-\textgreater9.1-\textgreater9.3-\textgreater9.2-\textgreater3.11-\textgreater3.6-\textgreater\\ 4.2-\textgreater3.10-\textgreater3.9-\textgreater3.1-\textgreater2.1\end{tabular}                                                                                                                                                                                                                                                                                                                                                                                                                                                                                                                                                                                                                                                                                                                                                                                                                                                                                                                                                                                                                                                            \\ \hline
			Test7 & 1549 & 3809.218 & 38725                        & 13.93\%                      & 9                               & 44                                 & \begin{tabular}[c]{@{}l@{}}1.0-\textgreater2.0-\textgreater3.0-\textgreater2.0-\textgreater3.0-\textgreater4.0-\textgreater\\ 3.0-\textgreater9.0-\textgreater3.0-\textgreater5.0-\textgreater7.0-\textgreater8.0-\textgreater\\ 7.0-\textgreater5.0-\textgreater6.0-\textgreater5.0-\textgreater3.0-\textgreater9.0-\textgreater\\ 3.0-\textgreater2.0-\textgreater1.0\end{tabular} & \begin{tabular}[c]{@{}l@{}}1.1-\textgreater2.1-\textgreater3.3-\textgreater3.2-\textgreater2.1-\textgreater3.9-\textgreater\\ 3.10-\textgreater3.5-\textgreater3.10-\textgreater3.5-\textgreater3.10-\textgreater3.4-\textgreater\\ 3.11-\textgreater3.7-\textgreater3.11-\textgreater3.10-\textgreater3.4-\textgreater3.6-\textgreater\\ 4.1-\textgreater4.3-\textgreater3.6-\textgreater3.11-\textgreater9.2-\textgreater9.1-\textgreater\\ 9.2-\textgreater3.11-\textgreater5.7-\textgreater5.4-\textgreater5.3-\textgreater5.2-\textgreater\\ 7.1-\textgreater7.5-\textgreater7.4-\textgreater7.6-\textgreater7.1-\textgreater7.7-\textgreater\\ 7.4-\textgreater7.7-\textgreater8.1-\textgreater8.5-\textgreater8.4-\textgreater8.6-\textgreater\\ 8.2-\textgreater8.8-\textgreater8.4-\textgreater8.10-\textgreater8.11-\textgreater8.3-\textgreater\\ 8.5-\textgreater7.1-\textgreater6.2-\textgreater6.1-\textgreater6.5-\textgreater6.9-\textgreater\\ 6.6-\textgreater6.5-\textgreater6.8-\textgreater6.4-\textgreater5.3-\textgreater5.10-\textgreater\\ 3.11-\textgreater9.1-\textgreater9.3-\textgreater9.2-\textgreater3.11-\textgreater3.10-\textgreater\\ 3.5-\textgreater3.10-\textgreater3.9-\textgreater2.1-\textgreater2.2-\textgreater1.1\end{tabular}\\
			\hline
		\end{tabular}%
	}
\end{table}

\end{document}